\documentclass[10pt,twocolumn,letterpaper]{article}

\usepackage[pagenumbers]{cvpr} 

\usepackage[dvipsnames]{xcolor}

\usepackage{graphicx}
\usepackage{amsmath}
\usepackage{amssymb}
\usepackage{booktabs}

\definecolor{cvprblue}{rgb}{0.21,0.49,0.74}
\usepackage[pagebackref,breaklinks,colorlinks,citecolor=cvprblue]{hyperref}

\usepackage[capitalize]{cleveref}
\crefname{section}{Sec.}{Secs.}
\Crefname{section}{Section}{Sections}
\Crefname{table}{Table}{Tables}
\crefname{table}{Tab.}{Tabs.}

\def\confName{CVPR}
\def\confYear{2024}

\usepackage{color}
\usepackage{xcolor}
\definecolor{deepGreen}{RGB}{0,153,0}
\definecolor{orange}{RGB}{255,125,0}

\def\GTgreen#1{\textcolor[RGB]{0,255,0}{#1}}

\definecolor{sainone}{RGB}{236, 242, 249}
\definecolor{saintwo}{RGB}{255, 230, 204}

\newcommand{\keypoint}[1]{\vspace{0.1cm}\noindent\textbf{#1}\;}
\newcommand{\cut}[1]{}
\usepackage{multirow}
\usepackage{amsmath}
\usepackage{amssymb}
\usepackage{mathtools}
\usepackage{arydshln}
\usepackage{soul}
\usepackage{nicefrac}
\usepackage{booktabs}
\usepackage{stmaryrd}
\usepackage{caption}
\usepackage{graphicx}
\usepackage{colortbl}
\definecolor{gray}{gray}{0.9}
\definecolor{pink}{RGB}{255, 234, 232}
\sethlcolor{pink}
\usepackage[accsupp]{axessibility}
\usepackage{scalerel,graphicx,xparse}

\newcommand{\MYhref}[3][blue]{\href{#2}{\color{#1}{#3}}}

\makeatletter
\apptocmd\@maketitle{{\myfigure{}\par}}{}{}
\makeatother
\usepackage{capt-of,etoolbox}

\makeatletter
\newcommand\notsotiny{\@setfontsize\notsotiny\@vipt\@viipt}
\makeatother

\usepackage{xcolor,pifont}
\newcommand*\colourcheck[1]{%
  \expandafter\newcommand\csname #1check\endcsname{\textcolor{#1}{\ding{52}}}%
}
\newcommand*\colourcross[1]{%
  \expandafter\newcommand\csname #1cross\endcsname{\textcolor{#1}{\ding{55}}}%
}
\colourcheck{blue}
\colourcheck{green}
\colourcheck{black}
\colourcross{red}
\colourcross{black}

\title{\vspace{-0.8cm}It's All About \textit{Your} Sketch: Democratising Sketch Control in Diffusion Models \vspace{-0.7cm}}

\author{\MYhref[cvprblue]{https://subhadeepkoley.github.io}{Subhadeep Koley}\textsuperscript{1,2} \hspace{.2cm} \MYhref[cvprblue]{https://ayankumarbhunia.github.io}{Ayan Kumar Bhunia}\textsuperscript{1} \hspace{.2cm} \MYhref[cvprblue]{https://scholar.google.com/citations?user=SoQ1vtAAAAAJ}{Deeptanshu Sekhri}\textsuperscript{1} \hspace{.2cm} \MYhref[cvprblue]{https://aneeshan95.github.io}{Aneeshan Sain}\textsuperscript{1} \\  \MYhref[cvprblue]{https://www.pinakinathc.me}{Pinaki Nath Chowdhury}\textsuperscript{1} \hspace{.2cm} \MYhref[cvprblue]{https://www.surrey.ac.uk/people/tao-xiang}{Tao Xiang}\textsuperscript{1,2} \hspace{.2cm} \MYhref[cvprblue]{https://www.surrey.ac.uk/people/yi-zhe-song}{Yi-Zhe Song}\textsuperscript{1,2} \\
\textsuperscript{1}SketchX, CVSSP, University of Surrey, United Kingdom.  \\
\textsuperscript{2}iFlyTek-Surrey Joint Research Centre on Artificial Intelligence.\\
{\tt\small \{s.koley, a.bhunia, d.sekhri, a.sain, p.chowdhury, t.xiang, y.song\}@surrey.ac.uk}\\
\small\url{https://subhadeepkoley.github.io/StableSketching}
}

\newcommand\myfigure{
\centering
\vspace{-0.9cm}
\captionsetup{type=figure} 
    \includegraphics[width=1.0\textwidth]{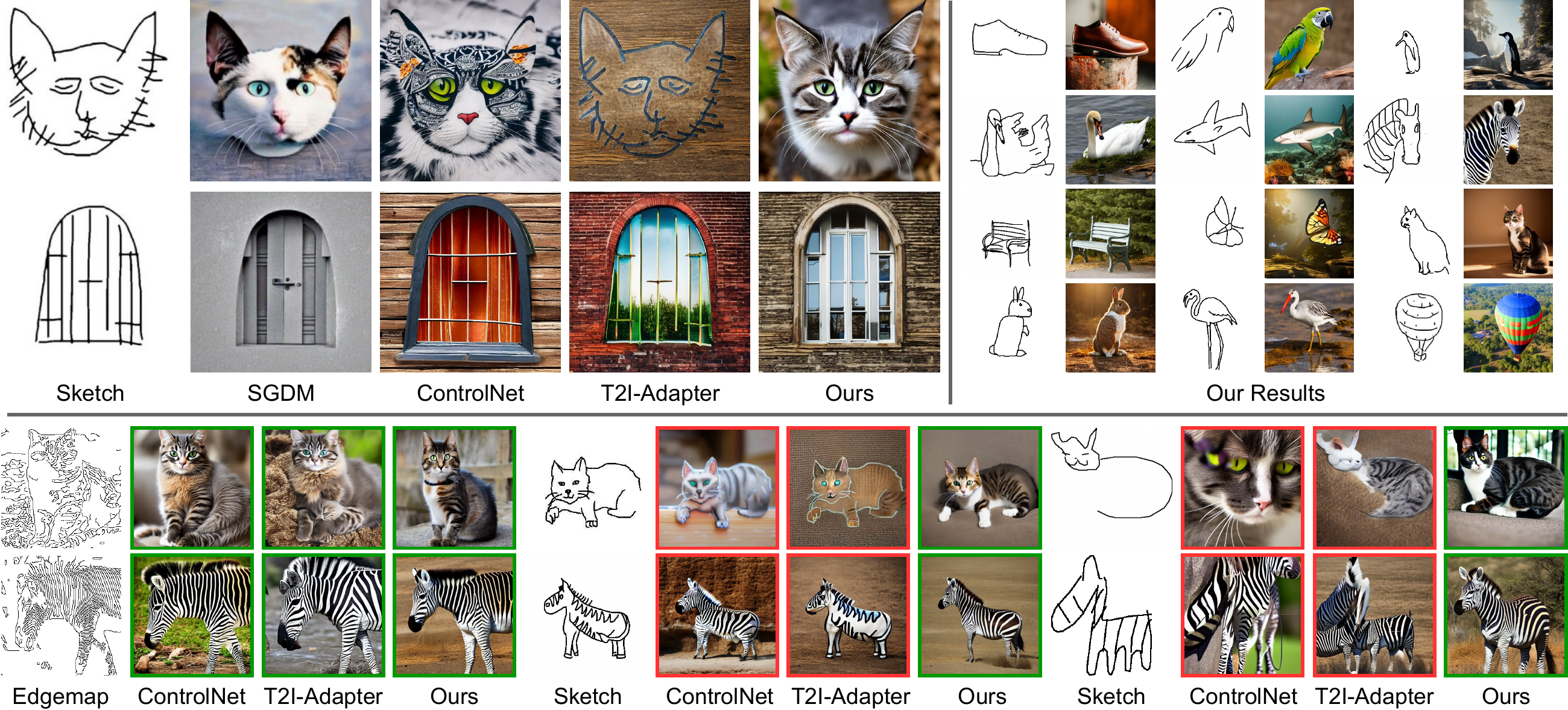}
    \vspace{-0.6cm}
\captionof{figure}{\textit{Top-left}: Comparison of images generated by our method with SGDM~\cite{voynov2023sketch}, ControlNet~\cite{zhang2023adding}, and T2I-Adapter~\cite{mou2023t2i}. \textit{Top-right}: A set of photos generated by our method. \textit{Bottom}: While existing methods~\cite{zhang2023adding, mou2023t2i} generate realistic images from \textit{pixel-perfect edgemaps}, they perform sub-optimally for \textit{freehand abstract sketches}. \textit{(Best view when zoomed in.)}
}

\label{fig:teaser}
\vspace{+0.3cm}
}

\begin{document}
\maketitle
\begin{abstract}
\vspace{-0.4cm}

This paper unravels the potential of sketches for diffusion models, addressing the deceptive promise of direct sketch control in generative AI. We importantly democratise the process, enabling amateur sketches to generate precise images, living up to the commitment of ``what you sketch is what you get''. A pilot study underscores the necessity, revealing that deformities in existing models stem from spatial-conditioning. To rectify this, we propose an abstraction-aware framework, utilising a sketch adapter, adaptive time-step sampling, and discriminative guidance from a pre-trained fine-grained sketch-based image retrieval model, working synergistically to reinforce fine-grained sketch-photo association. Our approach operates seamlessly during inference without the need for textual prompts; a simple, rough sketch akin to what you and I can create suffices! We welcome everyone to examine results presented in the paper and its supplementary. Contributions include democratising sketch control, introducing an abstraction-aware framework, and leveraging discriminative guidance, validated through extensive experiments.
\end{abstract}

\vspace{-1.4cm}
\section{Introduction}
This paper is dedicated to unlocking the full potential of \textit{your} sketches to control diffusion models \cite{rombach2022high, ho2022classifier, ho2020denoising}. Diffusion models ~\cite{rombach2022high, ho2022classifier, ho2020denoising, dhariwal2021diffusion} have made a significant impact, empowering individuals to unleash their visual creativity -- consider prompts like ``astronauts riding a horse on Mars" and other ``creative'' ones of your own! While prevailing in text-to-image generation~\cite{dhariwal2021diffusion, saharia2022photorealistic, rombach2022high}, recent works~\cite{voynov2023sketch, zhang2023adding, mou2023t2i} have started to question the expressive power of text as a conditioning modality. This shift has led to an exploration of sketches -- a modality that offers a degree of fine-grained control that is unparalleled by text~\cite{chowdhury2023scenetrilogy, sangkloy2022sketch}, resulting in generated content of closer resemblance. The promise is ``what you sketch is what you get''.

This promise is, however, deceptive. Current works (\eg, ControlNet~\cite{zhang2023adding}, T2I-Adapter~\cite{mou2023t2i}) predominantly focus on curated edgemap-like sketches -- you better sketch like a trained artist, otherwise ``what you get'' will literally be reflecting deformities captured in your (``half-decent'') sketch (\cref{fig:teaser}). The primary goal of this paper is to democratise sketch control in diffusion models, empowering real amateur sketches to generate photo-precise images, ensuring that ``what you get'' aligns with your \textit{intended sketch}, regardless of how well you drew it!  To achieve this, we draw insights from the sketch community~\cite{sain2021stylemeup, yang2022finding, sain2023exploiting, koley2024how, koley2023you} and introduce, for the first time, an awareness of sketch abstraction (as a result of varying drawing skills) into the generative process. This novel approach permits sketches of different abstraction levels to guide the generation process while maintaining output fidelity.

\begin{figure*}[!htbp]
    \centering
    \includegraphics[width=0.95\linewidth]{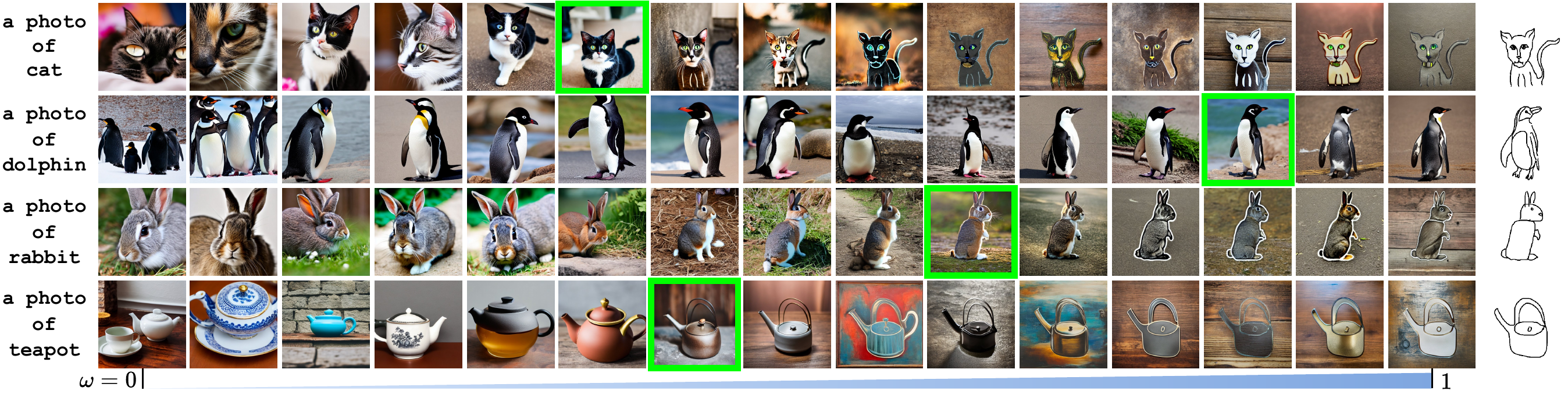}
    \vspace{-0.3cm}
    \caption{Images generated by T2I-Adapter~\cite{mou2023t2i} for different sketch-guidance factors ($\omega\in [0,1]$). Determining the optimum $\omega$ to obtain an ideal balance (\GTgreen{green}-bordered) between \textit{photorealism} and \textit{sketch-fidelity} requires manual intervention and is sample-specific. A high value of $\omega$ works well for less deformed sketches, while the same for an abstract sketch produces deformed outputs and vice-versa.}
    \label{fig:pilot}
    \vspace{-0.3cm}
\end{figure*}

We conduct a pilot study to reaffirm the necessity of our research (\cref{sec:pilot}). In which, we identify that the deformed output of existing sketch-conditional diffusion models stems from their \textit{spatial-conditioning} approach -- they directly translate sketch contours into the output photo domain, therefore producing deformed output. Conventional means of controlling the influence of spatial sketch-conditioning on the final output via weighing factors~\cite{mou2023t2i, voynov2023sketch} or sampling tricks~\cite{zhang2023adding}, however, require careful tuning. Reducing output deformity by assigning less weight to the sketch-conditioning often makes the output more coherent with the textual description, thus reducing its fidelity to the guiding sketch; yet, assigning higher weight to the textual prompt introduces lexical ambiguity~\cite{schwartz2023discriminative}. On the contrary, avoiding lexical ambiguity by assigning a higher weight to the guiding sketch almost always produces deformed and non-photorealistic outputs~\cite{zhang2023adding, mou2023t2i, voynov2023sketch}. Last but not least, the sweet spot between the conditioning weights is different for different sketch instances (as seen in \cref{fig:pilot}).

As such, our goal is to craft an effective sketch-conditioning strategy that not only operates without \textit{any} textual prompts during inference but is also \textit{abstraction-aware}. At the core of our work is a sketch adapter that transforms an input sketch into its \textit{equivalent textual embedding}, directing the denoising process of the diffusion model via cross-attention. Through the use of a smart time-step sampling strategy, we ensure the adaptability of the denoising process to the abstraction level of the input sketch. Additionally, by capitalising on the pre-trained knowledge of an off-the-shelf~\cite{sain2023clip} fine-grained sketch-based image retrieval (FG-SBIR) model, we incorporate discriminative guidance into our system for fine-grained sketch-photo association. Unlike widely used external classifier-guidance~\cite{dhariwal2021diffusion}, our proposed discriminative guidance mechanism does not require any specifically trained classifier capable of classifying \textit{both} noisy and real data. Lastly, even though our inference pipeline \textit{does not} rely on textual prompts, we use synthetically generated textual prompts during training to learn the sketch adapter with the limited sketch-photo paired data.

Our contributions are: \textit{(i)} we democratise sketch control, enabling real amateur sketches to generate accurate images, fulfilling the promise of ``what you sketch is what you get''. \textit{(ii)} we introduce an abstraction-aware framework that overcomes limitations of text prompts and spatial-conditioning. \textit{(iii)} we leverage discriminative guidance through a pre-trained FG-SBIR model for fine-grained sketch-fidelity. Extensive experiments validate the effectiveness of our method in addressing existing limitations in this domain.

\vspace{-0.2cm}
\section{Related Works}
\vspace{-0.2cm}
\keypoint{Diffusion Models for Vision Tasks.\ } Diffusion models \cite{ho2020denoising, ho2022classifier, sohl2015deep} have now become the gold-standard for different controllable image generation frameworks like DALL-E \cite{ramesh2021zero}, Imagen \cite{saharia2022photorealistic}, T2I-Adapter \cite{mou2023t2i}, ControlNet \cite{zhang2023adding}, etc.\cut{It performs iterative denoising via a denoising UNet~\cite{ronneberger2015u} on $2D$ Gaussian noise maps to generate high-resolution photorealistic images~\cite{ho2020denoising}. Based on the theory of physical diffusion process~\cite{sohl2015deep, song2019generative}, the diffusion model converts a clean image $\mathbf{x}_0$ into a noisy image $\mathbf{x}_T$ by iteratively adding random Gaussian noise to $\mathbf{x}_0$ for $T$ time-steps~\cite{ho2020denoising}. The opposite procedure (during inference) involves recovering $\mathbf{x}_0$ from $\mathbf{x}_T$ with iterative denoising~\cite{ho2020denoising}.} Besides image generation, several methods like Dreambooth~\cite{ruiz2023dreambooth}, Imagic~\cite{kawar2023imagic}, Prompt-to-Prompt~\cite{hertz2022prompt}, SDEdit~\cite{meng2021sdedit}, SKED~\cite{mikaeili2023sked} extend it for realistic image editing. Beyond image generation and editing, diffusion model is also used in several downstream vision tasks like recognition~\cite{li2023your}, semantic~\cite{baranchuk2021label} and panoptic~\cite{xu2023open} segmentation, image-to-image translation~\cite{tumanyan2023plug}, medical imaging~\cite{de2023medical}, image correspondence~\cite{tang2023emergent}, image retrieval \cite{koley2024text}, etc.

\keypoint{Sketch for Visual Content Creation.} Following its success in sketch-based image retrieval (SBIR)~\cite{sain2023clip, chowdhury2022partially, bhunia2021more}, sketches are now being used in other downstream tasks like saliency detection~\cite{bhunia2023sketch2saliency}, augmented reality~\cite{luo2022structure, luo20233d}, medical image analysis~\cite{kobayashi2023sketch}, object detection~\cite{chowdhury2023what}, class-incremental learning~\cite{bhunia2022doodle}, etc. Apart from the plethora of sketch-based $2D$ and $3D$ image generation and editing frameworks~\cite{koley2023picture, liu2020unsupervised, richardson2021encoding, ham2022cogs, zhang2023adding, mou2023t2i, voynov2023sketch, mikaeili2023sked, wang2022pretraining}, sketches are also getting significant traction in other visual content creation tasks like animation generation~\cite{smith2023method} and inbetweening~\cite{shen2023bridging}, garment design~\cite{li2018foldsketch, chowdhury2022garment}, caricature generation \cite{chen2023democaricature}, CAD modelling~\cite{li2022free2cad, yu2022piecewise}, anime editing~\cite{huang2023anifacedrawing}, etc.

\keypoint{Sketch-to-Image (S2I) Generation.}
Prior GAN-based S2I models typically leverage either contextual loss~\cite{lu2018image}, multi-stage generation~\cite{ghosh2019interactive}, etc.\ or performs latent mapping~\cite{koley2023picture, richardson2021encoding} on top of pre-trained GANs. Among diffusion-based frameworks, PITI~\cite{wang2022pretraining} trains a dedicated encoder to map the guiding sketch to the pre-trained diffusion model's latent manifold, SDEdit~\cite{meng2021sdedit} sequentially adds noise to the guiding sketch and iteratively denoise it based on a text prompt, while SGDM~\cite{voynov2023sketch} trains an MLP that maps the latent feature of the noisy images to the guiding sketches in order to force the intermediate noisy images to closely follow the guidance sketches. Among more recent multi-conditional (\eg, depth map, colour palate, key pose, etc.) frameworks, ControlNet~\cite{zhang2023adding} learns to control a frozen diffusion model by creating a trainable copy of its UNet encoders and connects it with the frozen model with \textit{zero-convolution}~\cite{zhang2023adding}, while T2I-Adapter~\cite{mou2023t2i} learns an encoder to extract features from the guidance signal (\eg, sketch) and conditions the generation process by adding the guidance features with the intermediate UNet features at each scale.
While existing methods can generate photorealistic images from precise edgemaps, they struggle with abstract freehand sketches (see Fig.~\ref{fig:teaser}). Furthermore, it is noteworthy that almost all of the diffusion-based S2I models~\cite{voynov2023sketch, wang2022pretraining, zhang2023adding, mou2023t2i, meng2021sdedit} rely heavily on highly-engineered and detailed textual prompts.

\vspace{-0.25cm}
\section{Revisiting Diffusion Model (DM)}
\vspace{-0.2cm}
\keypoint{Overview.} Diffusion models comprises two complementary random processes \textit{viz.\ } \textit{``forward''} and \textit{``reverse''}~\cite{ho2020denoising} diffusion. Forward diffusion process iteratively adds Gaussian noise of varying magnitude to a clean training image $\mathbf{x}_0 \in \mathbb{R}^{h\times w\times 3}$ for {$t$} time-steps to yield a noisy image $\mathbf{x}_t \in \mathbb{R}^{h\times w\times 3}$ as:

\vspace{-0.5cm}
\begin{equation}
    \mathbf{x}_t = \sqrt{\bar{\alpha}_t}\mathbf{x}_{0} + (\sqrt{1-\bar{\alpha}_t})\epsilon
    \vspace{-0.1cm}
    \label{eq:noising}
\end{equation}

\noindent where, $\epsilon$$\sim$$\mathcal{N}(0,\mathbf{I})$, $t$$\sim$${U}(0,T)$, and $\{\alpha_t\}_1^T$ is a pre-defined noise schedule with $\bar{\alpha}_t = \prod_{i=1}^{t} \alpha_i$~\cite{ho2020denoising}. Reverse diffusion process trains a modified denoising UNet~\cite{ronneberger2015u} $\mathcal{F}_\theta (\cdot)$, that estimates the input noise $\epsilon \approx  \mathcal{F}_\theta(\mathbf{x}_t,t)$ from the noisy image $\mathbf{x}_t$ at each time-step $t$. $\mathcal{F}_\theta$ being trained with an {$l_2$ loss}~\cite{ho2020denoising} can reverse the effect of the forward diffusion procedure. {During inference, starting} from a random $2D$ noise $\mathbf{x}_T$ sampled from a Gaussian distribution, $\mathcal{F}_\theta$ is applied iteratively (for $T$ time-steps) to denoise $\mathbf{x}_t$ at each time-step $t$ to get a cleaner image $\mathbf{x}_{t-1}$, eventually leading to a cleanest image $\mathbf{x}_0$ of the original target distribution~\cite{ho2020denoising}.

The unconditional denoising diffusion process could be made ``conditional’’ by influencing the $\mathcal{F}_\theta$ with auxiliary conditioning signals ${d}$ (\eg, textual description \cite{rombach2022high, ramesh2022hierarchical, saharia2022photorealistic}, etc.). Thus, $\mathcal{F}_\theta(\mathbf{x}_t,t,{d})$ could perform denoising on $\mathbf{x}_t$ while being guided by ${d}$ via cross-attention~\cite{rombach2022high}.

\label{ldm}
\keypoint{Latent Diffusion Model.} Unlike standard diffusion models \cite{dhariwal2021diffusion, ho2020denoising}, \textit{Latent Diffusion Model} \cite{rombach2022high} (\textit{a.k.a.} Stable Diffusion--SD) performs denoising diffusion on the latent space for faster and more stable training \cite{rombach2022high}. SD first \textit{trains an autoencoder} {(consists of an encoder $\mathcal{E}(\cdot)$ and a decoder $\mathcal{D}(\cdot)$ in series)} to convert the input image $\mathbf{x}_0 \in \mathbb{R}^{h\times w\times 3}$ to its latent representation $\mathbf{z}_0 =\mathcal{E}(\mathbf{x}_0)\in \mathbb{R}^{\frac{h}{8}\times \frac{w}{8}\times d}$. Later, SD \textit{trains a modified denoising UNet} \cite{ronneberger2015u} $\epsilon_\theta(\cdot)$ to perform denoising directly on the latent space. The textual prompt ${d}$ upon passing through a CLIP textual encoder \cite{radford2021learning} $\mathbf{T}(\cdot)$ produces the corresponding token-sequence that influences the intermediate feature maps of the UNet via cross-attention \cite{rombach2022high}. SD trains with an $l_2$ loss as:

\vspace{-0.2cm}
\begin{equation}
    \mathcal{L}_{\text{SD}} = \mathbb{E}_{\mathbf{z}_t,t,{d},\epsilon} ({||\epsilon-\epsilon_{\theta}(\mathbf{z}_t,t,\mathbf{T}(d))||}_2^2)
    \label{eq:sd}
\end{equation}

During inference, SD \textit{discards} $\mathcal{E}(\cdot)$, directly sampling a noisy latent $\mathbf{z}_T$ from a Gaussian distribution~\cite{rombach2022high}. It then  estimates noise from $\mathbf{z}_T$ iteratively for $T$ iterations via $\epsilon_\theta$ (conditioned on $d$) to obtain a clean latent $\Hat{\mathbf{z}}_0$. The frozen decoder generates the final image as: $\Hat{\mathbf{x}}_0=\mathcal{D}(\Hat{\mathbf{z}}_0)$~\cite{rombach2022high}.

\vspace{-0.2cm}
\section{What's wrong with Sketch-to-Image DM}
\vspace{-0.1cm}
\label{sec:pilot}

Recent controllable image generation methods like ControlNet \cite{zhang2023adding}, T2I-Adapter \cite{mou2023t2i}, etc.\ offer extreme photorealism, supporting different conditioning signals (\eg, depth map, label mask, edgemap, etc.) However, conditioning the same from sparse \textit{freehand sketches} is often sub-optimal (\cref{fig:teaser}).

\vspace{0.1cm}
\noindent \textbf{Sketch \vs Other Conditional Inputs.}
Sparse and binary {freehand sketches} while good for providing \textit{fine-grained} spatial cues~\cite{yu2016sketch, chowdhury2023what, bhunia2023sketch2saliency}, often depict significant shape-deformity~\cite{hertzmann2020line, eitz2012humans, sain2021stylemeup} and hold far less contextual information~\cite{tumanyan2023plug} than other \textit{pixel-perfect} conditioning signals like depth maps, normal maps, or pixel-level segmentation masks. Hence, conditioning from freehand sketches is non-trivial and needs to be handled uniquely unlike the rest of the pixel-perfect conditioning signals.

\begin{figure}[!htbp]
    \centering
    \includegraphics[width=1\linewidth]{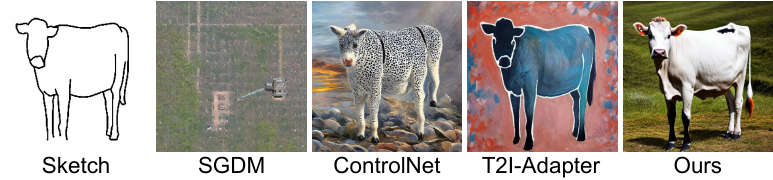}
    \vspace{-0.7cm}
    \caption{Passing null prompt (\ie, $\mathtt{``~"}$) in existing~\cite{voynov2023sketch, zhang2023adding, mou2023t2i} sketch-conditioned DMs significantly distorts the output quality.}
    \label{fig:no_prompt}
    \vspace{-0.4cm}
\end{figure}

\noindent \textbf{Sketch \vs Text Conditioning: A Trade-off.} Previous S2I diffusion models~\cite{mou2023t2i, zhang2023adding, voynov2023sketch} exhibit two major challenges. \textit{Firstly}, quality of generated outputs being highly dependent on precise and accurate textual prompts~\cite{zhang2023adding}, inconsistencies or lack of suitable prompts can negatively impact (\cref{fig:no_prompt}) the results~\cite{mou2023t2i, zhang2023adding}. \textit{Secondly}, ensuring a balance between the influence of sketch and text-conditioning on the final output requires manual intervention, which can be challenging. Adjusting the weighting of these factors often results in a \textit{trade-off} between output's coherence with the text and fidelity to the sketch~\cite{mou2023t2i}. In some cases, giving higher weight to text can lead to lexical ambiguity~\cite{schwartz2023discriminative}, while prioritising sketch tends to produce distorted and non-photorealistic results~\cite{mou2023t2i, voynov2023sketch}. Achieving photorealistic output from existing S2I DMs~\cite{mou2023t2i, voynov2023sketch} thus demands \textit{meticulous fine-tuning} of these weights, where the optimal balance varies for different sketch instances as seen in \cref{fig:pilot}.

\vspace{-0.25cm}

\vspace{0.1cm}
\noindent \textbf{Problems with Spatial-Conditioning for Sketches.} We identify that the deformed and non-photorealistic (\eg, edge-bleeding in \cref{fig:pilot}) outputs of existing sketch-conditional DMs \cite{mou2023t2i, zhang2023adding, voynov2023sketch} are primarily a consequence of their \textit{spatial-conditioning} approach. T2I-Adapter \cite{mou2023t2i} directly integrates the \textit{spatial features} of the conditioning-sketch into the UNet encoder's feature maps, while ControlNet \cite{zhang2023adding} applies this to skip connections and middle blocks. SGDM \cite{voynov2023sketch}, on the other hand, projects the latent features of noisy images to \textit{spatial} edgemaps guiding the denoising process towards following the edgemaps. Additionally, these models are trained and tested with \textit{synthetically-generated}~\cite{su2021pixel, canny1986computational, xie2015holistically} edgemaps/contours rather than \textit{real} freehand sketches. Instead, we aim to devise an effective conditioning strategy for \textit{real} freehand sketches while ensuring that the output faithfully captures an end-users' \textit{semantic intent} \cite{koley2023picture} without any deformities.

\vspace{-0.2cm}
\section{Proposed Methodology}
\begin{figure}[!t]
    \centering
    \includegraphics[width=1\linewidth]{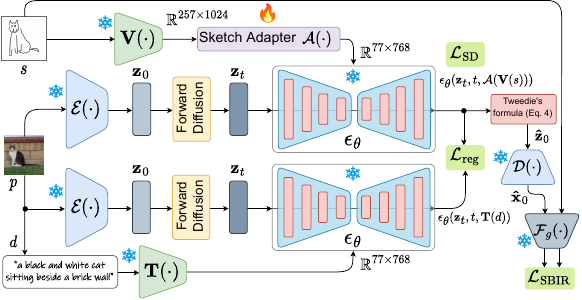}
    \vspace{-0.75cm}
    \caption{Our overall training pipeline. \textit{(More in the text.)}}
    \label{fig:arch}
    \vspace{-0.6cm}
\end{figure}

\vspace{-0.2cm}
\keypoint{Overview.}  We aim to eliminate \textit{spatial sketch-conditioning} by converting the input sketch into an \textit{equivalent {fine-grained} textual embedding}, thereby preserving users' semantic-intent without \textit{pixel-level} spatial alignment. Consequently, our method would alleviate issues pertaining to spatial distortions (\eg, deformed shapes, edge-bleeding, etc.) while maintaining \textit{fine-grained fidelity} to the input sketch. We introduce three salient designs (\cref{fig:arch}) -- \textit{(i)} fine-grained discriminative loss for maintaining the \textit{fine-grained} sketch-photo correspondence (\cref{sec:fine}). \textit{(ii)} guiding our training process with textual prompts (\textit{not} used during inference), as a means of \textit{super-concept} preservation (\cref{sec:super}). Finally, \textit{(iii)} unlike the \textit{uniform} time-step ($t$) sampling of prior arts~\cite{zhang2023adding, voynov2023sketch}, we introduce a \textit{sketch-abstraction-aware} $t$-sampling (\cref{sec:imp}). For a highly abstract sketch, a higher probability is assigned to larger $t$ and vice-versa.

\vspace{-0.1cm}
\subsection{Sketch Adapter}
\vspace{-0.1cm}
\label{sec:adapter}
Aiming to mitigate the evident disadvantages (\cref{sec:pilot}) of direct \textit{spatial-conditioning} approach of existing sketch-conditional diffusion models (\eg, ControlNet~\cite{zhang2023adding}, T2I-Adapter~\cite{mou2023t2i}, etc.), we take a parallel approach to ``sketch-condition'' the generation process via cross-attention. In that, instead of treating the input sketches \textit{spatially}, we encode them as a sequence of feature vectors~\cite{image_variation} as an \textit{equivalent {fine-grained} textual embedding}. Direct spatial-conditioning enforces the model to \textit{remember} the contextual information rather than \textit{understanding} it~\cite{yang2023paint}. This results in a direct translation of the strong sketch features (\eg, stroke boundaries) into the output photo. To overcome this, we aim to increase the hardness of the problem by compressing the spatial sketch input to a \textit{bottlenecked-representation} via sketch adapter.

In particular, given a sketch $s$, we use a pre-trained CLIP~\cite{radford2021learning} ViT-L/14 image encoder $\mathbf{V}(\cdot)$ to generate its patch-wise sketch embedding $\mathbf{s}=\mathbf{V}(s)\in \mathbb{R}^{257\times 1024}$. Our sketch adapter $\mathcal{A}(\cdot)$ consists of $1$-dimensional convolutional and vanilla attention~\cite{vaswani2017attention} modules followed by FC layers. The convolutional and FC layers handle the dimension mismatch between text and sketch-embedding (\ie, $\mathbb{R}^{257\times 1024}$ $\rightarrow$ $\mathbb{R}^{77\times 768}$), whereas the attention module tackles the large sketch-text domain gap. The patch-wise sketch embedding $\mathbf{s}$ upon passing through $\mathcal{A}(\cdot)$ generates the equivalent textual embedding as $\mathbf{\Hat{s}}=\mathcal{A}(\mathbf{s})\in\mathbb{R}^{77\times 768}$. Now replacing the textual conditioning in \cref{eq:sd} with our sketch adapter conditioning, the modified loss objective becomes:

\vspace{-0.6cm}
\begin{equation}
    \mathcal{L}_{\text{SD}} = \mathbb{E}_{\mathbf{z}_t,t,{s},\epsilon} ({||\epsilon-\epsilon_{\theta}(\mathbf{z}_t,t,\mathcal{A}(\mathbf{V}(s)))||}_2^2)
    \vspace{-0.1cm}
    \label{eq:sd_modified}
\end{equation}

{Once trained, the sketch adapter efficiently converts an input sketch $s$ into its {equivalent textual embedding} $\hat{\mathbf{s}}$, which through cross-attention controls the denoising process of SD~\cite{rombach2022high}.} Nonetheless, conditioning solely via the proposed sketch adapter poses multiple challenges -- \textit{(i)} sparse freehand sketches and pixel-perfect photos depict a huge domain gap. The standard $l_2$ loss~\cite{rombach2022high} of a text-to-image diffusion model is not enough to ensure a \textit{fine-grained} matching between sketch and photo. \textit{(ii)} training a robust sketch adapter from the \textit{limited} available sketch-photo pairs is difficult. Consequently, during training, we aim to use \textit{pseudo texts} as a learning signal to guide the training of our sketch adapter. Please note, our inference pipeline \textit{does not} involve any textual prompts. \textit{(iii)} the sketch adapter treats \textit{all} sketch samples \textit{equally} regardless of their \textit{abstraction levels}. While this equal treatment might suffice for dense pixel-level conditioning, it might be inadequate for sparse sketches, as different sketches depicting different abstraction levels are not \textit{semantically-equal} \cite{bhunia2022sketching, yang2021sketchaa}.

\vspace{-0.1cm}
\subsection{Fine-Grained Discriminative Learning}
\vspace{-0.1cm}
\label{sec:fine}
To ensure a \textit{fine-grained matching} between sparse freehand sketches and pixel-perfect photos, we utilise a pre-trained fine-grained (FG) SBIR  model \cite{sain2023clip} $\mathcal{F}_g(\cdot)$. A photo sits close to its paired sketch in a pre-trained FG-SBIR model's \textit{discriminative} latent embedding space compared to other unpaired ones \cite{sain2023clip}. Previous attempts at guiding the diffusion process with external discriminative models include classifier-guidance~\cite{dhariwal2021diffusion} that require a pre-trained fixed-class classifier capable of classifying \textit{both} noisy and real data~\cite{dhariwal2021diffusion} {to guide the denoising procedure~\cite{dhariwal2021diffusion}}. However, as our frozen FG-SBIR model is not trained on \textit{noisy} data, {it requires a \textit{clean} image at each $t$, to perform in an \textit{off-the-shelf} manner. Now, for each $t$, as the denoiser estimates that noise $\epsilon_t\approx\epsilon_\theta(\mathbf{z}_t, t, \mathcal{A}(\mathbf{V}(s)))$, which was added to $\mathbf{z}_0$ to get $\mathbf{z}_t$ during forward diffusion, we can use \cref{eq:noising} to recreate $\mathbf{z}_0$ from $\epsilon_t$. Specifically, we utilise Tweedie's formula \cite{kim2021noise2score} to estimate~\cite{yang2023paint, avrahami2022blended, kwon2022diffusion} the clean latent image $\Hat{\mathbf{z}}_0$ from the $t^\text{th}$-step noisy latent $\mathbf{z}_t$ in a single-step for efficient training as:}

\vspace{-0.4cm}
\begin{equation}
    \Hat{\mathbf{z}}_0(\mathbf{z}_t) := \frac{\mathbf{z}_t-\sqrt{1-\bar{\alpha}_t}~\epsilon_\theta(\mathbf{z}_t, t, \mathcal{A}(\mathbf{V}(s)))}{\sqrt{\bar{\alpha}_t}}
    \vspace{-0.2cm}
\end{equation}

$\Hat{\mathbf{z}}_0$ upon passing through SD's~\cite{rombach2022high} frozen VAE decoder $\mathcal{D}(\cdot)$ approximates the clean image $\Hat{\mathbf{x}}_0$ (\cref{ldm}). To learn the sketch adapter $\mathcal{A}$, we use a discriminative SBIR loss that calculates cosine similarity $\delta(\cdot,\cdot)$ between $s$ and $\Hat{\mathbf{x}}_0$ as:

\vspace{-0.2cm}
\begin{equation}
\label{eq:fgsbir_loss}
    \mathcal{L}_{\text{SBIR}} = 1 - \delta\left({\mathcal{F}_g(s) \cdot \mathcal{F}_g(\Hat{\mathbf{x}}_0)}\right)
    \vspace{-0.1cm}
\end{equation}

\subsection{Super-concept Preservation Loss}
\vspace{-0.1cm}
\label{sec:super}
An inherent complementarity exists between \textit{sketch} and \textit{text} \cite{chowdhury2023scenetrilogy}. A textual caption of an image can correspond to multiple \textit{plausible} photos in the embedding space. Adding a sketch \textit{with it} however, narrows down the scope to a \textit{particular} image \cite{chowdhury2023scenetrilogy, sangkloy2022sketch} (\ie, fine-grained). We posit that a textual description being less fine-grained than a sketch \cite{chowdhury2023scenetrilogy, song2017fine, yang2023paint}, acts as a \textit{super-concept} of the corresponding sketch. Although we \textit{do not} use any textual prompt during inference, we aim to use them during training of our sketch adapter. Text-to-image diffusion models being trained on a large corpus of text-image pairs~\cite{rombach2022high}, implicitly hold superior text-to-image generation capability (although \textit{not} fine-grained \cite{ge2023expressive}). We thus aim to use this super-concept knowledge from textual descriptions to distil the large-scale text-to-image knowledge of a pre-trained SD to train our sketch-adapter with \textit{limited} sketch-photo paired data.

As our sketch-photo ($s,p$) dataset \cite{sangkloy2016sketchy} lacks paired textual captions, we use a pre-trained state-of-the-art image captioner \cite{li2022blip} to synthetically generate caption $d$ for every ground truth photo $p$. Then, at each $t$, the noise predicted through \textit{text-conditioning} $(\mathbf{T}(d))$ acts as a reference to calculate a regularisation loss to learn the sketch adapter $\mathcal{A}$ as:

\vspace{-0.6cm}
\begin{equation}
\mathcal{L}_{\text{reg}} = ||\epsilon_\theta(\mathbf{z}_t,t,\mathbf{T}(d))-\epsilon_\theta(\mathbf{z}_t,t,\mathcal{A}(\mathbf{V}(s)))||_2^2
\vspace{-0.1cm}
\end{equation}

\subsection{Abstraction-aware Importance Sampling}
\label{sec:imp}
\begin{figure}[!htbp]
    \centering
    \includegraphics[width=1\linewidth]{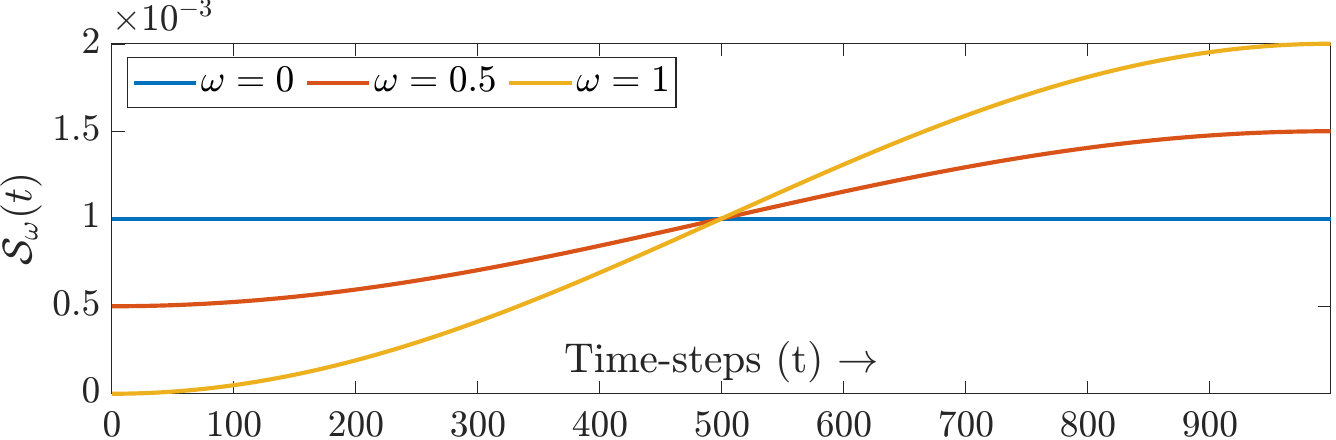}
    \vspace{-0.7cm}
    \caption{Abstraction-aware $t$-sampling function for different $\omega$.}
    \label{fig:sampling}
    \vspace{-0.4cm}
\end{figure}

\vspace{-0.1cm}
Existing literature~\cite{huang2023reversion,mou2023t2i,huang2023collaborative, yang2023paint} indicates that during the denoising process, high-level semantic structures of the output image tend to manifest in the early stages, while finer appearance details emerge later. Synthetic pixel-perfect conditioning signals (\eg, depth map \cite{ranftl2022towards}, key pose \cite{cao2019openpose}, edgemap \cite{canny1986computational}, etc.) exhibit minimal subjective abstraction~\cite{hertzmann2020line}. In contrast, human-drawn freehand sketches exhibit varying abstraction levels, influenced by factors like skill, style, and subjective interpretation~\cite{sain2021stylemeup, sain2023exploiting}. Thus, uniform time-step sampling~\cite{huang2023reversion} for abstract sketches may compromise output generation quality and sketch-fidelity. Hence, we propose adjusting the time-step sampling procedure based on the input sketch's abstraction level~\cite{yang2022finding}. For highly abstract sketches, we skew the sampling distribution to emphasise the later $t$ values that govern the high-level semantics in the output. Instead of sampling the time-step from uniform distribution $t$$\sim$${U}(0,T)$, we sample from:

\vspace{-0.2cm}
\begin{equation}
    \mathcal{S}_\omega(t)=\frac{1}{T}\left(1-\omega~\mathrm{cos}\frac{\pi t}{T}\right)
    \vspace{-0.1cm}
    \label{eq:samplig}
\end{equation}

\noindent where, $\mathcal{S}_\omega(\cdot)$ is our \textit{abstraction-aware $t$-sampling function}, where increasing or decreasing $\omega\in(0,1]$, controls the skewness of this sampling probability density function. Pushing $\omega$ towards $1$ increases the probability of sampling a larger $t$ value (\cref{fig:sampling}). We aim to make this skewness-controlling $\omega$ value sketch-abstraction specific.

\vspace{-0.3cm}

Now the question remains as to how we can quantify the abstraction level of a freehand sketch. Taking inspiration from \cite{yang2022finding}, we design a CLIP \cite{radford2021learning}-based (a generic classifier) sketch classifier with a MagFace \cite{meng2021magface}-based loss where the $l_2$-norm of a sketch feature $\mathbf{a}\in[0,1]$, denotes how closely it sits from its respective class-centre. While $\mathbf{a}$ $\rightarrow$ $1$ represents edgemap-like \textit{less abstract} sketches, $\mathbf{a}$ $\rightarrow$ $0$ denotes \textit{highly-abstract and deformed} ones. We posit that edgemaps being less deformed (\ie, easier to classify), will implicitly stay close to their respective class centres in the latent space. Whereas, freehand sketches being highly abstract and deformed (\ie, harder to classify), will be placed away from their corresponding class centres. We thus train the sketch classifier with sketches and synthesised \cite{chan2022drawings} edgemaps of the associated photos from Sketchy \cite{sangkloy2016sketchy}, using our classification loss:

\vspace{-0.3cm}
\begin{equation}
\footnotesize
    \mathcal{L}_{\text{abs}} = -\mathrm{log}\frac{e^{s~\mathrm{cos}(\theta_{y_i}+m(\mathbf{s}_i))}}{e^{s~\mathrm{cos}(\theta_{y_i}+m(\mathbf{s}_i))}+\sum_{j\neq y_i} e^{s~\mathrm{cos}~\theta_j}}+\lambda_gg(\mathbf{s}_i)
\end{equation}

\noindent where $s$ is a global scalar value, $\theta_{y_i}$ is the cosine similarity between extracted global visual feature (from CLIP~\cite{radford2021learning} visual encoder) of the $i^\text{th}$ sketch sample $\mathbf{s}_i=\mathbf{V}(s_i)\in\mathbb{R}^d$ with $l_2$-normalisation, and $j^\text{th}$ class centre $w_j\in\mathbb{R}^d$ computed from ground truth class labels by CLIP~\cite{radford2021learning} text encoder. $m(\mathbf{s}_i)$ is the magnitude-aware margin parameter $m(\mathbf{s}_i)=\frac{(u_m-l_m)}{(u_a-l_a)l_a+l_m}$, where $l_m$, $u_m$ denotes the lower and upper bounds of the margin, and $l_a$, $u_a$ denotes that of the feature magnitude. $g(\mathbf{s}_i)$ is a hyper-parameter ($\lambda_g$)-controlled regularisation term (see \cite{meng2021magface} for more details). With the trained classifier, given a sketch $s$, the \emph{scalar abstraction score} $\mathbf{a}\in[0,1]$ is given by the $l_2$-norm of the extracted sketch feature $\mathbf{V}(s)$. To keep parity with $\omega$, we complement $\mathbf{a}$ to get the sketch instance-specific  $\omega \leftarrow (1-\mathbf{a})$, followed by empirically clipping $\omega$ in the range $[0.2,0.8]$.

In summary, we train the sketch adapter $\mathcal{A}(\cdot)$ using sketch-abstraction-aware $t$-sampling with a total loss of $\mathcal{L}_\text{total}$=$\lambda_1\mathcal{L}_\text{SD}$+$\lambda_2\mathcal{L}_\text{SBIR}$+$\lambda_3\mathcal{L}_\text{reg}$. During inference, we compute the abstraction score of the input sketch, taking $l_2$-norm of classifier feature. Based on the abstraction level, we perform $t$-sampling. The input sketch passing through $\mathcal{A}$ controls the diffusion procedure and generates the output.

\vspace{-0.2cm}
\section{Experiments}
\vspace{-0.3cm}
\keypoint{Dataset and Implementation Details.\ } We train and evaluate our model on the Sketchy dataset \cite{sangkloy2016sketchy} containing $12,500$ images from $125$ categories with at least $5$ sketches per image with \textit{fine-grained} association. For training and evaluation, we split this dataset in $90$:$10$. We use Stable Diffusion v1.5 \cite{rombach2022high} in all experiments with a CLIP~\cite{radford2021learning} embedding dimension $d=768$. The sketch adapter is trained with a learning rate of $10^{-4}$, keeping the SD model, FG-SBIR backbone, and CLIP encoders frozen. We train our model for $50$ epochs using AdamW \cite{loshchilov2019decoupled} optimiser with $0.09$ weight decay, and batch size of $8$. Values of $\lambda_{1,2,3}$ are set to $1$, $0.5$, and $0.1$, empirically.

\vspace{-1mm}
\keypoint{Evaluation Metrics.} Following \cite{mou2023t2i, zhang2023adding, koley2023picture}, we quantitatively evaluate the generation quality and sketch-fidelity with four metrics -- \textit{Frech\`et Inception Distance-InceptionV3 (FID-I) \cite{karras2019style} and \textit{CLIP} (FID-C)} \cite{kynkaanniemi2022role} calculates the similarity between generated and real images using pre-trained InceptionV3 \cite{szegedy2016rethinking} and CLIP \cite{radford2021learning} ViT-B/32 models respectively. Lower values of FID-I and FID-C depict better generation quality. We measure the output image's fidelity to the input sketch using \textit{Fine-Grained Metric (FGM)} \cite{koley2023picture} which computes the cosine similarity between them via a pre-trained FG-SBIR model \cite{sain2023clip}, where a higher value denotes better fine-grained correspondence. Additionally, we also perform a human study to collect \textit{Mean Opinion Score (MOS)} \cite{huynh2010study}. Here, we asked $25$ \textit{non-artist} users to draw $40$ sketches each, and rate the generated photos on a discrete scale (interval=$0.5$) of $[1,5]$ (worst to best) based on \textit{output photorealism} and \textit{sketch-fidelity}. For each method, we compute the final MOS by averaging all its $1000$ MOS values.

\begin{figure*}[!htbp]
    \centering
    \includegraphics[width=0.95\linewidth]{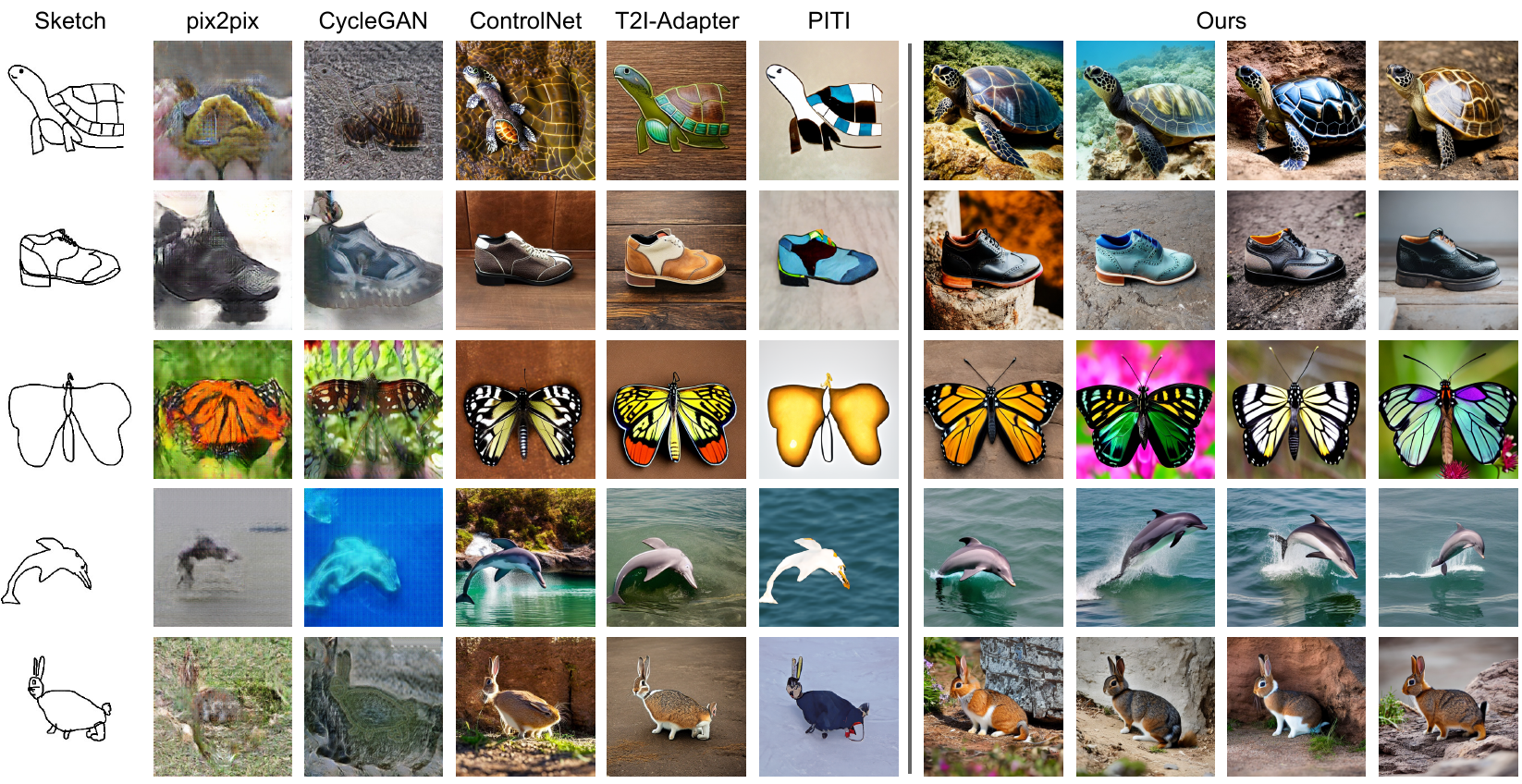}
    \vspace{-0.35cm}
    \caption{Qualitative comparison with SOTA sketch-to-image generation models on Sketchy~\cite{sangkloy2016sketchy}. For ControlNet~\cite{zhang2023adding}, T2I-Adapter~\cite{mou2023t2i}, and PITI~\cite{wang2022pretraining}, we use the fixed prompt $\mathtt{``a~photo~of~[CLASS]"}$, with $\mathtt{[CLASS]}$ replaced with corresponding class-labels of the input sketches.}
    \label{fig:qual}
    \vspace{-0.5cm}
\end{figure*}

\vspace{-0.5mm}
\keypoint{Competitors.} We compare against different diffusion and GAN-based state-of-the-art (SOTA) S2I models and two baselines. \textit{(i) Sketch-only \underline{B}aselines:} To alleviate the necessity of text, \textbf{B-Classification} first trains a prompt learning-based sketch classifier \cite{khattak2023maple} that classifies every sketch into one of the predefined classes. From predicted class labels, it forms a textual prompt (\ie, $\mathtt{``a~photo~of~[CLASS]"}$) to generate images using a frozen text-to-image SD model \cite{rombach2022high}. Given the input sketches, \textbf{B-Captioning} first generates detailed captions using a pre-trained image captioner \cite{li2022blip} from their paired photos, which are then used to generate images from a frozen SD model \cite{rombach2022high}. \textit{(ii) SOTAs:} Among diffusion-based SOTAs, we compare with \textbf{ControlNet} \cite{zhang2023adding}, \textbf{T2I-Adapter} \cite{mou2023t2i}, \textbf{SGDM} \cite{voynov2023sketch}, and \textbf{PITI} \cite{wang2022pretraining}. We also compare qualitatively against two GAN-based S2I paradigms \textit{viz.} \textbf{Pix2Pix} \cite{isola2017image} and \textbf{CycleGAN} \cite{zhu2017unpaired}. While we train ControlNet \cite{zhang2023adding}, T2I-Adapter \cite{mou2023t2i}, and PITI \cite{wang2022pretraining} on the entire Sketchy \cite{sangkloy2016sketchy} train set, we train pix2pix \cite{isola2017image}, and CycleGAN \cite{zhu2017unpaired} individually for each of the depicted classes (\cref{fig:qual}) from scratch with Sketchy \cite{sangkloy2016sketchy} sketch-photo pairs. We only perform a qualitative comparison with SGDM \cite{voynov2023sketch} by taking the results directly from the paper, as their model weights/code are unavailable. Notably, for diffusion-based SOTAs \cite{wang2022pretraining, zhang2023adding, mou2023t2i}, we use an \textit{additional} fixed textual prompt $\mathtt{``a~photo~of~[CLASS]"}$, replacing $\mathtt{[CLASS]}$ with class-labels of respective
input sketches.

\vspace{-0.1cm}
\subsection{Performance Analysis \& Discussion}
\vspace{-0.2cm}

\keypoint{Result Analysis.} Among GAN-based methods, pix2pix \cite{isola2017image} and CycleGAN \cite{zhu2017unpaired} depict visible deformities (\cref{fig:qual}) mostly due to their  weaker \cite{dhariwal2021diffusion} GAN-based generator, compared to an internet-scale pre-trained SD model \cite{rombach2022high}. Among diffusion-based SOTAs, although SGDM~\cite{voynov2023sketch} generates plausible colour schemes and styles, outputs exhibit substantial deformations (\cref{fig:teaser}). A similar observation can be made for PITI \cite{wang2022pretraining}, where generated images look non-photorealistic with pronounced edge-adherence (\cref{fig:qual}). Whereas, edge-bleeding (\cref{fig:qual}) is quite frequent for T2I-Adapter~\cite{mou2023t2i}. ControlNet~\cite{zhang2023adding} surpasses PITI \cite{wang2022pretraining}, SGDM~\cite{voynov2023sketch}, and T2I-Adapter \cite{mou2023t2i} in terms of photorealism but mostly follows the input sketch boundaries (\cref{fig:qual}). Contrarily, images generated by our method are more photorealistic with fewer deformities, capturing semantic-intent without transmitting edge boundaries in the output. Quantitative results presented in \cref{tab:benchmark} show B-Caption to surpass B-Classification (by $0.11$ FGM) thanks to the comparatively higher~\cite{li2022blip} generalisation potential of the captioning model~\cite{li2022blip} than the generic sketch classifier \cite{khattak2023maple}. Nonetheless, our method exceeds these baselines both in terms of generation quality and sketch-fidelity with an FID-C of $16.20$ and FGM of $0.81$. Due to its superior conditioning strategy, ControlNet \cite{zhang2023adding} achieves the lowest FID-I among all prior SOTAs (\cref{tab:benchmark}). Although less pronounced in terms of FID-I/FID-C, our method offers the highest \textit{fine-grained sketch-fidelity} with $23.45\%$ FGM improvement of over ControlNet \cite{zhang2023adding}. Finally, thanks to the photorealistic generation quality and fine-grained sketch correspondence, our method surpasses competitors in terms of MOS value from user-study with an average $1.36\pm 0.2$ point improvement. Notably, unlike ours, image generation via diffusion-based competitors needs textual prompts, the absence of which results in much worse output quality (\cref{fig:no_prompt}).

\vspace{-0.1cm}
\begin{table}[!htbp]
\renewcommand{\arraystretch}{0.9}
\setlength{\tabcolsep}{7pt}
\footnotesize
\centering
\caption{Benchmarks on the Sketchy~\cite{sangkloy2016sketchy} dataset.}
\vspace{-0.3cm}
\label{tab:benchmark}
\begin{tabular}{lcccc}
\toprule
\multicolumn{1}{c}{\multirow{2}{*}{Methods}} & \multirow{2}{*}{FID-I~$\downarrow$}  & \multirow{2}{*}{FID-C~$\downarrow$}  & \multirow{2}{*}{FGM~$\uparrow$} & MOS~$\uparrow$\\
& & & & $\mu \pm \sigma$\\
\cmidrule(lr){1-5}
ControlNet~\cite{zhang2023adding}       & 26.68 & 21.22 & 0.62 & 3.68$\pm$0.2  \\
T2I-Adapter~\cite{mou2023t2i}           & 26.94 & 18.92 & 0.56 & 3.11$\pm$0.6  \\
PITI~\cite{wang2022pretraining}         & 84.71 & 25.85 & 0.23 & 2.64$\pm$0.3 \\ \cmidrule(lr){1-5}
\cmidrule(lr){1-5}
B-Classification                        & 28.93 & 19.01 & 0.36 & 3.13$\pm$0.2  \\
B-Captioning                            & 28.31 & 18.81 & 0.47 & 3.21$\pm$0.4  \\
\rowcolor{YellowGreen!40}
\textbf{\textit{Proposed}}              & \bf25.07 & \bf16.20 & \bf0.81 & \bf4.52$\pm$0.1 \\ \bottomrule
\end{tabular}
\vspace{-0.25cm}
\end{table}

\begin{figure}[!htbp]
    \centering
    \includegraphics[width=1\linewidth]{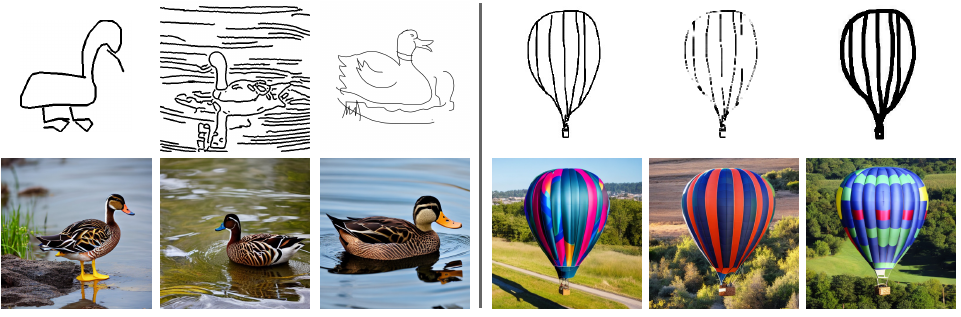}
    \vspace{-0.6cm}
    \caption{Examples showing generalisation potential across different \textit{datasets} (left) and \textit{stroke-styles} (right).}
    \label{fig:cross_dataset}
    \vspace{-0.35cm}
\end{figure}

\begin{figure}[!htbp]
    \centering
    \includegraphics[width=1\linewidth]{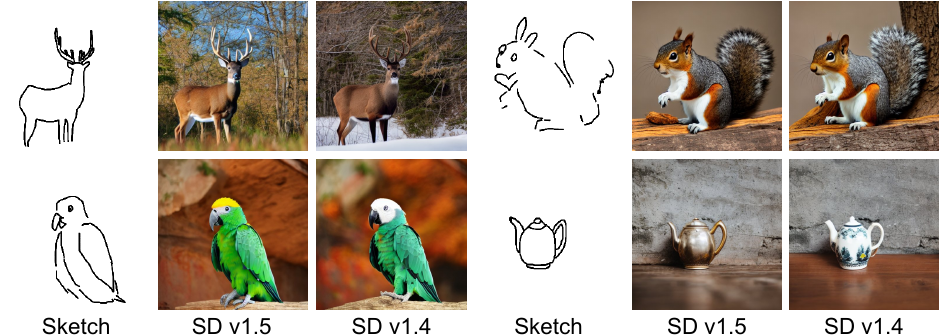}
    \vspace{-0.7cm}
    \caption{Illustration of \textit{cross-model} generalisation. Our method trained with SD v1.5~\cite{rombach2022high}, performs well on other unseen SD variants (\eg, v1.4) without further fine-tuning.}
    \label{fig:model_var}
    \vspace{-0.65cm}
\end{figure}

\keypoint{Generalisation Potential.} As our method alleviates the direct spatial influence of input sketches in the denoising process, it enables generalisation across multiple dimensions. \cref{fig:cross_dataset} shows that our sketch-adapter trained on Sketchy, generalises well on random sketch samples from TU-Berlin \cite{eitz2012humans} and QuickDraw \cite{ha2017neural} datasets, on synthetically generated \cite{canny1986computational} edgemaps, and to different stroke-styles. Furthermore, as our sketch adapter does not distort the original text-to-image pre-training of the frozen SD model, the same adapter could be used to perform sketch-conditional generation from other versions of the SD model (\cref{fig:model_var}).

\begin{figure}[!htbp]
    \centering
    \includegraphics[width=1\linewidth]{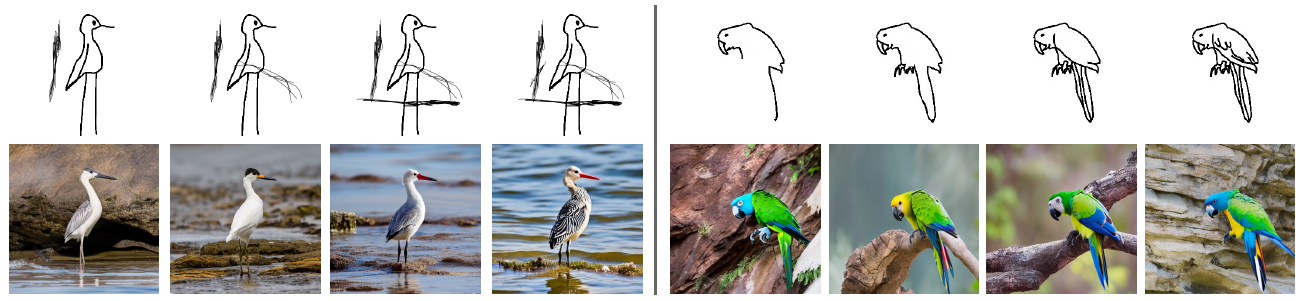}
    \vspace{-0.6cm}
    \caption{Examples depicting the effect of adding \textit{noisy strokes} (left) and generation from \textit{partially-completed} sketches (right).}
    \label{fig:noisy}
    \vspace{-0.2cm}
\end{figure}

\keypoint{Robustness and Sensitivity.}
Amateur freehand sketching often introduces irrelevant and noisy strokes~\cite{bhunia2022sketching}. We thus demonstrate our model's resilience to such strokes by progressively adding them during inference, and assessing its performance. On the other hand, to judge our model's stability against partially-complete sketches, we render input sketches at $\{25,50,75,100\}\%$ prior to generation. As our method is devoid of \textit{direct} spatial-conditioning, outputs remain relatively stable (\cref{fig:noisy}) even for spatially distorted sketches (\eg, noisy or partially-complete).

\vspace{-0.1cm}

\begin{figure}[t]
    \centering
    \includegraphics[width=1\linewidth]{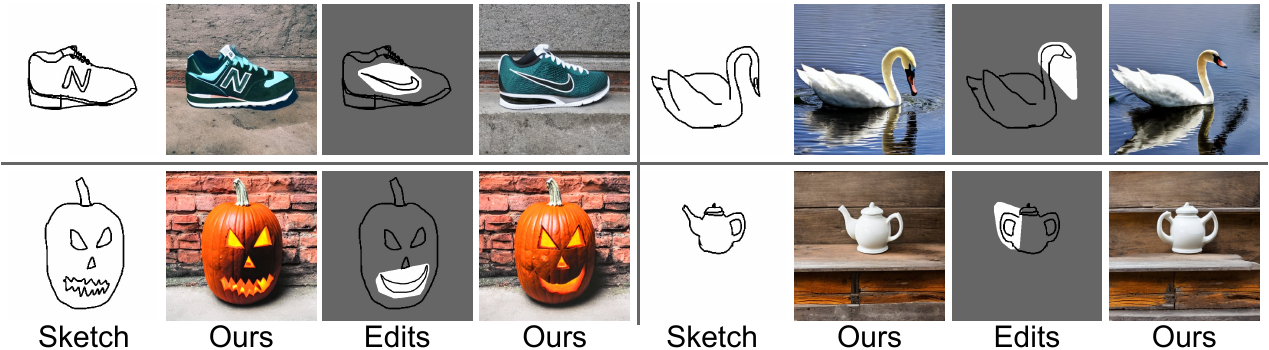}
    \vspace{-0.7cm}
    \caption{Our method seamlessly transfers local semantic edits on input sketches into output photos. \textit{(Best view when zoomed in.)}}
    \label{fig:editing}
    \vspace{-0.65cm}
\end{figure}

\keypoint{Fine-grained Semantic Editing.} Harnessing the large-scale pre-training of the frozen SD model~\cite{rombach2022high}, our method enables \textit{fine-grained} semantic editing. Here, fixing the generation seed, and performing local semantic edits in the sketch-domain produces seamless edited images (\cref{fig:editing}).

\vspace{-0.5cm}

\vspace{.3cm}
\subsection{Ablation on Design}
\vspace{-0.1cm}
\noindent{\textbf{[i] Importance of Sketch Adapter.}} Our sketch adapter (\cref{sec:adapter}) converts an input sketch to its corresponding \textit{ textual equivalent embedding}. To judge its efficacy, we replace it with simple convolutional and FC-layers converting the $\mathbb{R}^{257\times1024}$ sketch embedding to equivalent $\mathbb{R}^{77\times768}$ textual embedding. Although less pronounced in FID scores, the FGM score plummets substantially ($49.38\%$) in case of \textbf{w/o Sketch adapter} (\cref{tab:abal}), indicating the significance of the proposed adapter in maintaining high sketch-fidelity.

\noindent{\textbf{[ii] Why Discriminative Learning?}} Fine-grained discriminative loss (Eq.\ \ref{eq:fgsbir_loss}) helps the conditioning process by distilling knowledge learned inside a pre-trained FG-SBIR model. As seen in \cref{tab:abal}, a noticeable FGM drop ($44.44\%$) for \textbf{w/o Discriminative learning} indicates that fine-grained sketch-conditioning is incomplete without explicit discriminative learning via $\mathcal{L}_{\text{SBIR}}$.

\noindent{\textbf{[iii] Does Abstraction-aware Importance Sampling help?}} Unlike existing sketch-conditional DMs, we take freehand sketch abstraction into account via \textit{abstraction-aware} $t$-sampling. Omitting it results (\cref{tab:abal}) in a sharp increase in FID-I scores ($26.64\%$). We hypothesise that in absence of the proposed adaptive $t$-sampling, the system treats \textit{all} sketches \textit{equally}, regardless of their abstraction level, resulting in sub-optimal performance.

\begin{figure}[!htbp]
    \centering
    \includegraphics[width=0.95\linewidth]{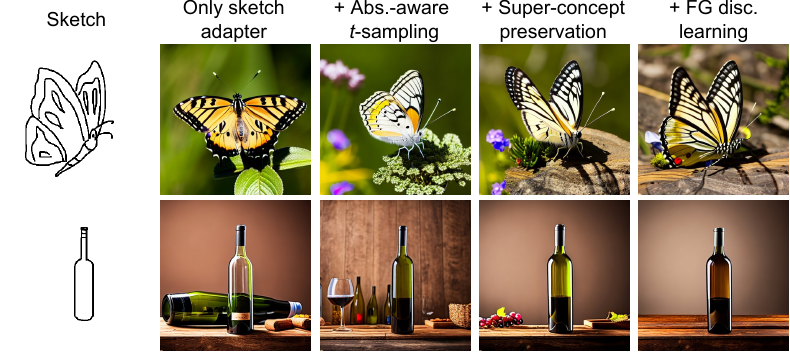}
    \vspace{-0.3cm}
    \caption{Visual ablation of different design components.}
    \label{fig:abal}
    \vspace{-0.2cm}
\end{figure}

\noindent{\textbf{[iv] Impact of Super-concept Preservation.}} Although our inference procedure \textit{does not} use any textual prompt, we employ them during our training process to facilitate the preservation of \textit{super-concepts}. Eliminating this again destabilises the system causing an additional $15.06\%$ and $17.28\%$ decline in FID-C and FGM scores (\cref{tab:abal}). This justifies our incorporation of synthetic text prompts during training, as it aligns well with the original text-to-image generation objective of the pre-trained SD model~\cite{rombach2022high}. Visual ablation results are presented in \cref{fig:abal}.

\vspace{-0.2cm}

\begin{table}[!htbp]
\renewcommand{\arraystretch}{1}
\setlength{\tabcolsep}{7pt}
\footnotesize
\centering
\caption{Ablation on design.}
\vspace{-0.3cm}
\label{tab:abal}
\begin{tabular}{lccc}
\toprule
\multicolumn{1}{c}{\multirow{1}{*}{Methods}} & \multirow{1}{*}{FID-I~$\downarrow$}  & \multirow{1}{*}{FID-C~$\downarrow$}  & \multirow{1}{*}{FGM~$\uparrow$}\\
\cmidrule(lr){1-4}
w/o Sketch adapter                & 29.23 & 20.34 & 0.41 \\
w/o Discriminative learning       & 29.14 & 19.97 & 0.45 \\
w/o Super-concept preservation    & 27.21 & 18.64 & 0.67 \\
w/o Abs.-aware $t$-sampling       & 31.75 & 23.17 & 0.55 \\ \cmidrule(lr){1-4}
Ours (SD v1.4)                    & 26.12 & 17.09 & 0.77 \\
\rowcolor{YellowGreen!40}
\textbf{\textit{Ours-full}}       & \bf25.07 & \bf16.20 & \bf0.81 \\ \bottomrule
\end{tabular}
\vspace{-0.2cm}
\end{table}

\subsection{Failure Cases \& Future Works}
\begin{figure}[!htbp]
    \centering
    \includegraphics[width=1\linewidth]{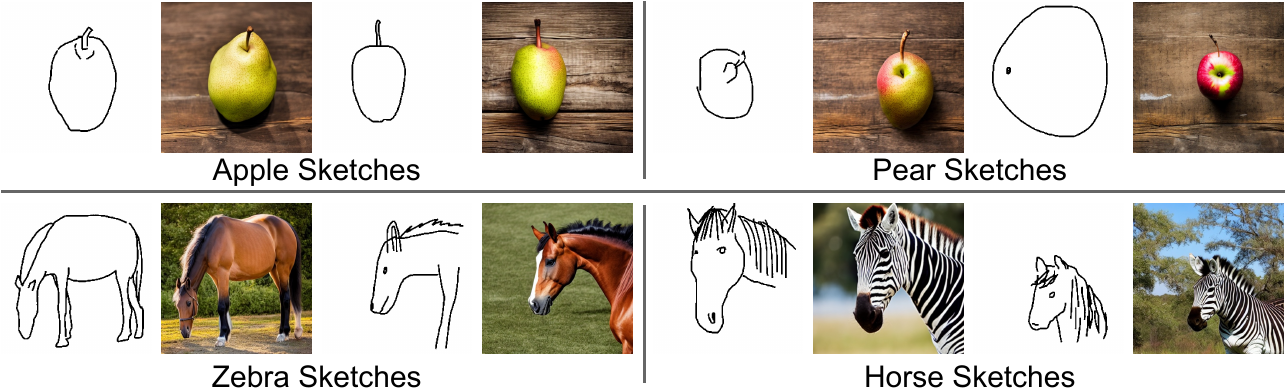}
    \vspace{-0.7cm}
    \caption{Failure cases where sketches from certain classes (\eg, zebra) produce images from other similar-looking classes (\eg, horse) or vice-versa. Please note that we \textit{do not} use text prompts.}
    \label{fig:failure}
    \vspace{-0.6cm}
\end{figure}

\vspace{-0.2cm}
Despite showcasing superior generation quality without significant deformations, our method has a few limitations. For Instance, it sometimes struggles to determine the correct class of the input due to
\textit{categorical-ambiguity}, especially when two different objects look very similar shape-wise (\cref{fig:failure}) in their \textit{abstract} and \textit{deformed} sketch forms (\eg, apple \vs pear, guitar \vs violin). {In future, we aim to extend our method with the flexibility to include additional class labels. The sketch+label \textit{composed-conditioning} \cite{saito2023pic2word} might mitigate the categorical-ambiguity of confusing classes.}

\vspace{-0.2cm}

\vspace{-0.1cm}
\section{Conclusion}
\vspace{-0.2cm}

Our work takes a significant step towards democratising sketch control in diffusion models. We exposed the limitations of current approaches, showcasing the deceptive promise of sketch-based generative AI. By introducing an abstraction-aware framework, featuring a sketch adapter, adaptive time-step sampling, and discriminative guidance, we empower amateur sketches to yield precise, high-fidelity images without the need for textual prompts during inference. We welcome the community to scrutinise our results. Please refer to the demo video for a detailed real-time comparison with state-of-the-arts.

{\small
\bibliographystyle{ieeenat_fullname}
\bibliography{arxiv}
}

\clearpage

\onecolumn{
\centering
\title{\Large{\textbf{Supplementary material for\\ It's All About \textit{Your} Sketch: Democratising Sketch Control in Diffusion Models}}}
\vspace{0.4cm}

\author{\MYhref[cvprblue]{https://subhadeepkoley.github.io}{Subhadeep Koley}\textsuperscript{1,2} \hspace{.2cm} \MYhref[cvprblue]{https://ayankumarbhunia.github.io}{Ayan Kumar Bhunia}\textsuperscript{1} \hspace{.2cm} \MYhref[cvprblue]{https://scholar.google.com/citations?user=SoQ1vtAAAAAJ}{Deeptanshu Sekhri}\textsuperscript{1} \hspace{.2cm} \MYhref[cvprblue]{https://aneeshan95.github.io}{Aneeshan Sain}\textsuperscript{1} \\  \MYhref[cvprblue]{https://www.pinakinathc.me}{Pinaki Nath Chowdhury}\textsuperscript{1} \hspace{.2cm} \MYhref[cvprblue]{https://www.surrey.ac.uk/people/tao-xiang}{Tao Xiang}\textsuperscript{1,2} \hspace{.2cm} \MYhref[cvprblue]{https://www.surrey.ac.uk/people/yi-zhe-song}{Yi-Zhe Song}\textsuperscript{1,2} \\
\textsuperscript{1}SketchX, CVSSP, University of Surrey, United Kingdom.  \\
\textsuperscript{2}iFlyTek-Surrey Joint Research Centre on Artificial Intelligence.\\
{\tt\small \{s.koley, a.bhunia, d.sekhri, a.sain, p.chowdhury, t.xiang, y.song\}@surrey.ac.uk}
}

}

\maketitle

\section*{A. Additional Qualitative Results}
\begin{figure}[!htbp]
    \centering
    \includegraphics[width=1\linewidth]{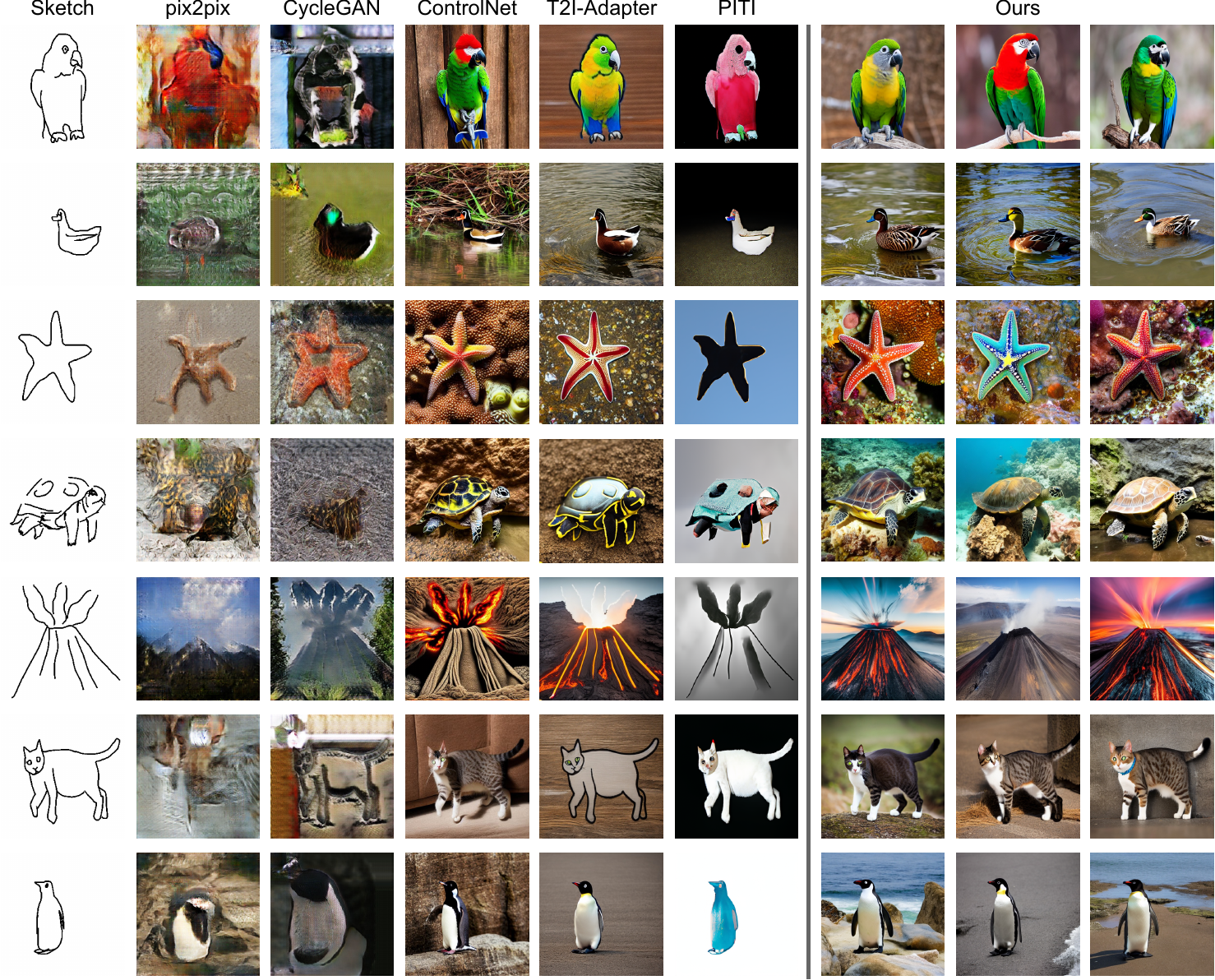}
    \vspace{-0.2cm}
    \caption{Qualitative comparison with SOTAs. For ControlNet \cite{zhang2023adding}, T2I-Adapter \cite{mou2023t2i}, and PITI \cite{wang2022pretraining}, we use the fixed prompt $\mathtt{``a~photo~of~[CLASS]"}$, with $\mathtt{[CLASS]}$ replaced with corresponding class-labels of the input sketches. \textit{(Best view when zoomed in.)}}
    \label{fig:comp_1}
\end{figure}

\begin{figure}[!htbp]
    \centering
    \includegraphics[width=1\linewidth]{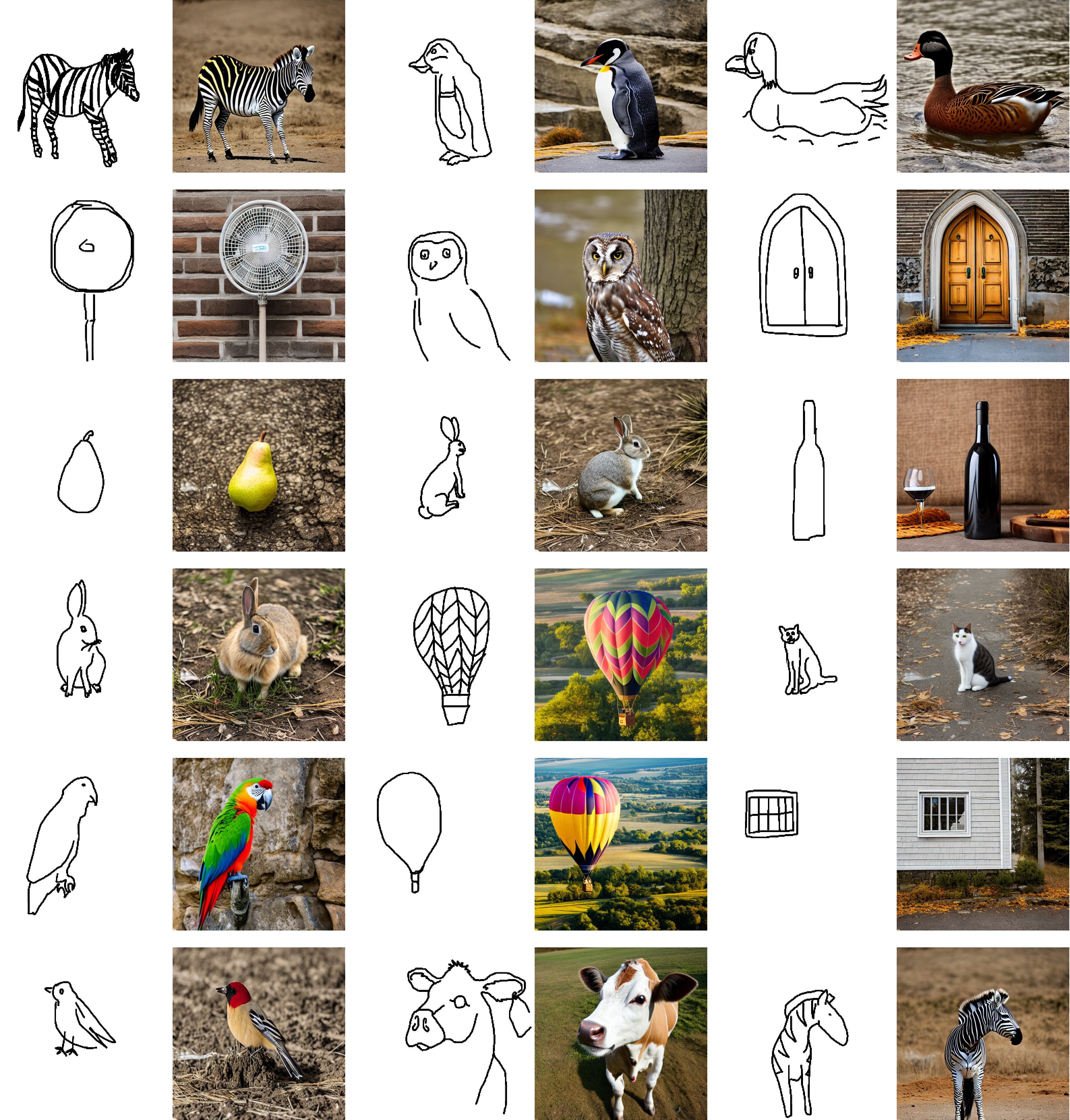}
    \vspace{-0.6cm}
    \caption{Images generated by our method.}
    \label{fig:qual_1}
\end{figure}

\begin{figure}[!htbp]
    \centering
    \includegraphics[width=1\linewidth]{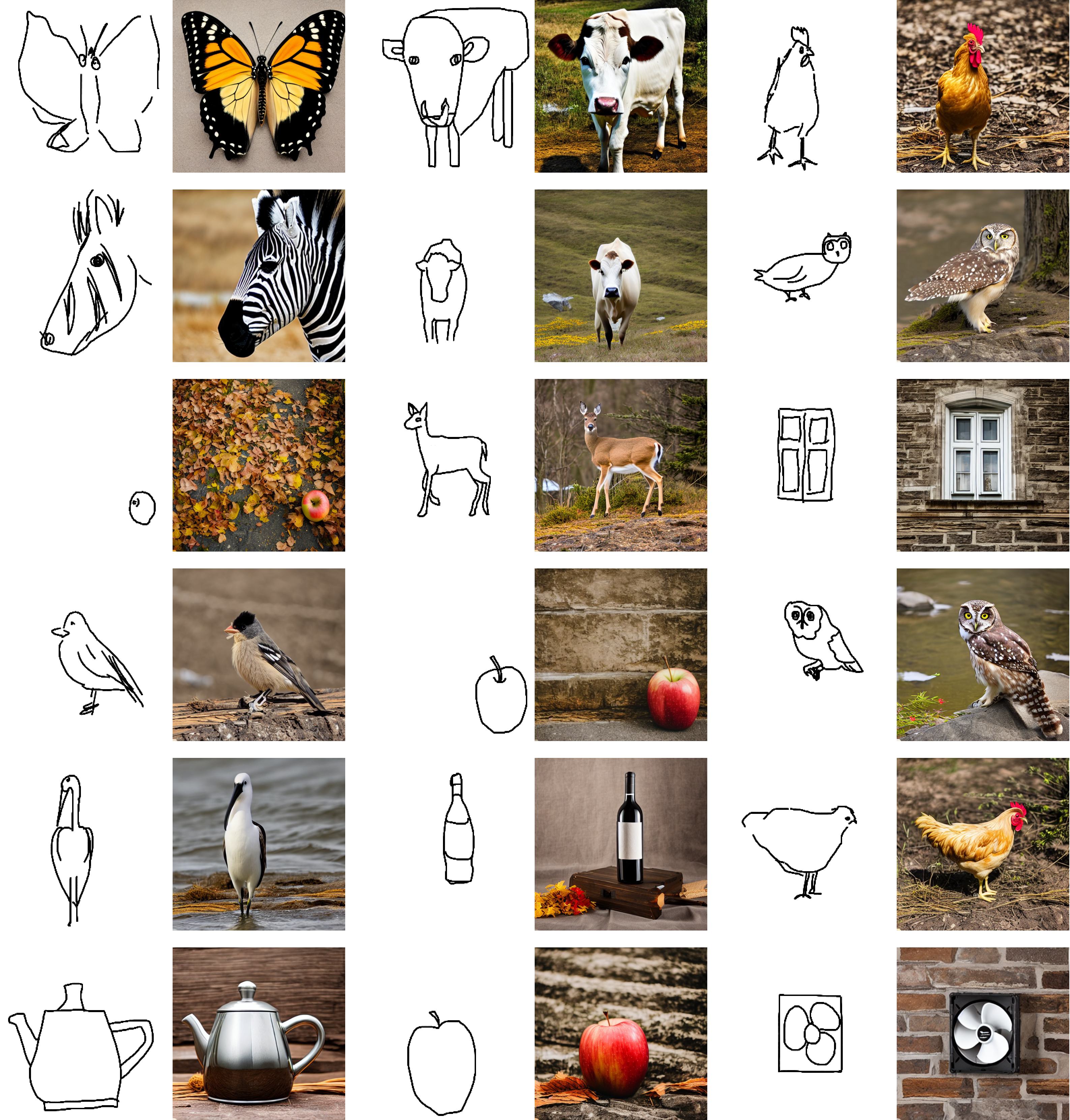}
    \vspace{-0.6cm}
    \caption{Images generated by our method.}
    \label{fig:qual_2}
\end{figure}

\begin{figure}[!htbp]
    \centering
    \includegraphics[width=1\linewidth]{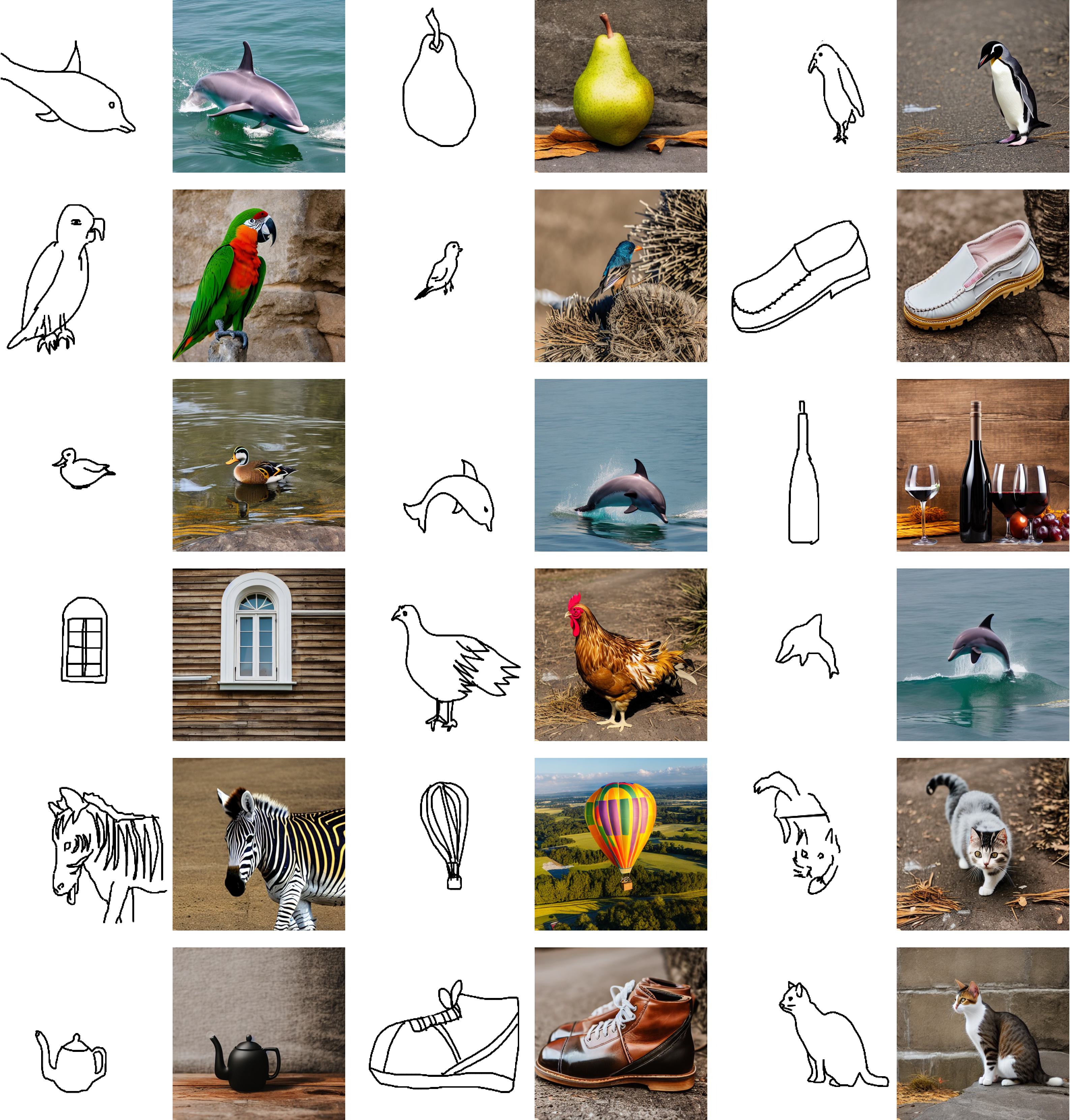}
    \vspace{-0.6cm}
    \caption{Images generated by our method.}
    \label{fig:qual_3}
\end{figure}

\begin{figure}[!htbp]
    \centering
    \includegraphics[width=1\linewidth]{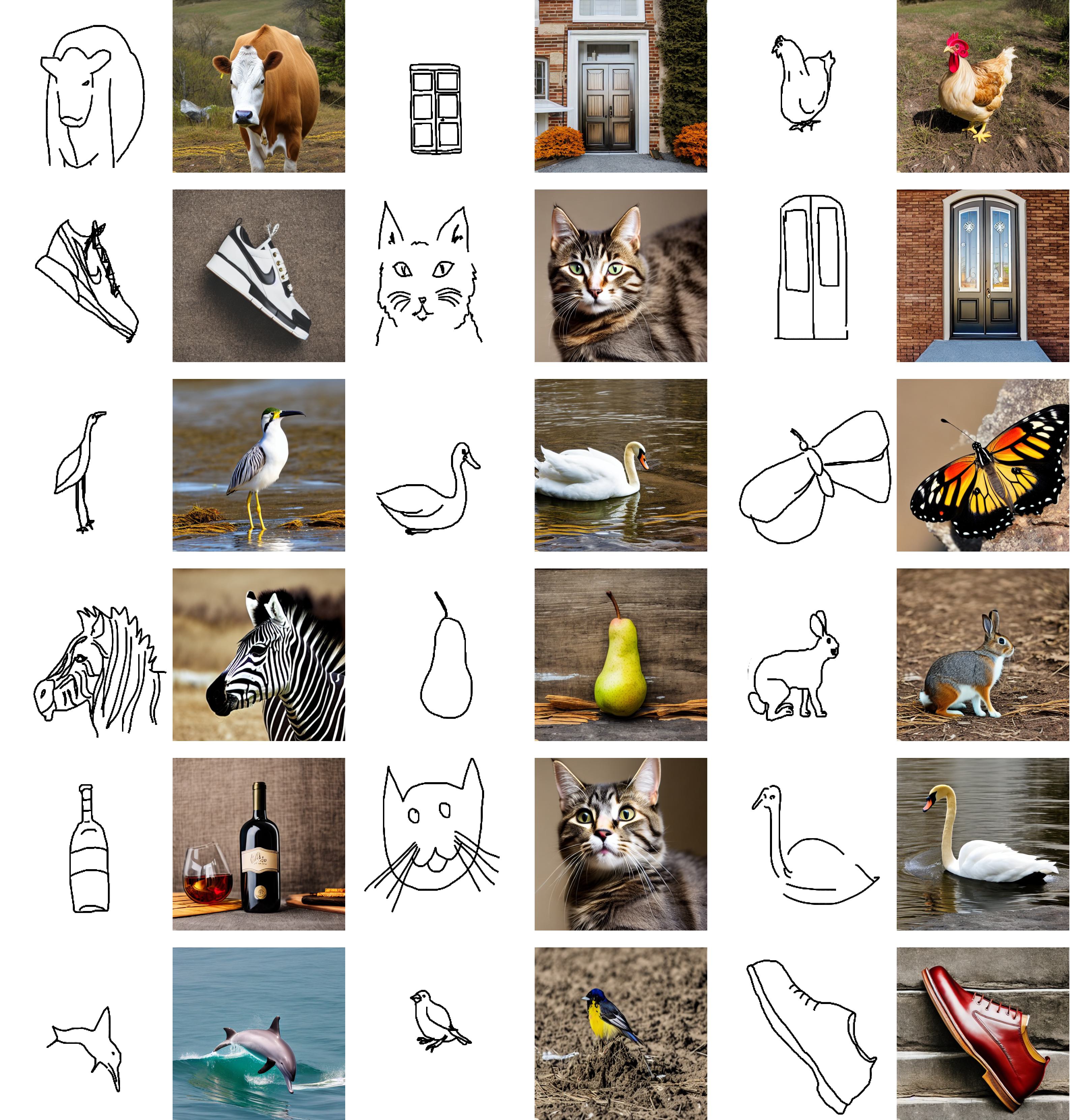}
    \vspace{-0.6cm}
    \caption{Images generated by our method.}
    \label{fig:qual_4}
\end{figure}

\begin{figure}[!htbp]
    \centering
    \includegraphics[width=1\linewidth]{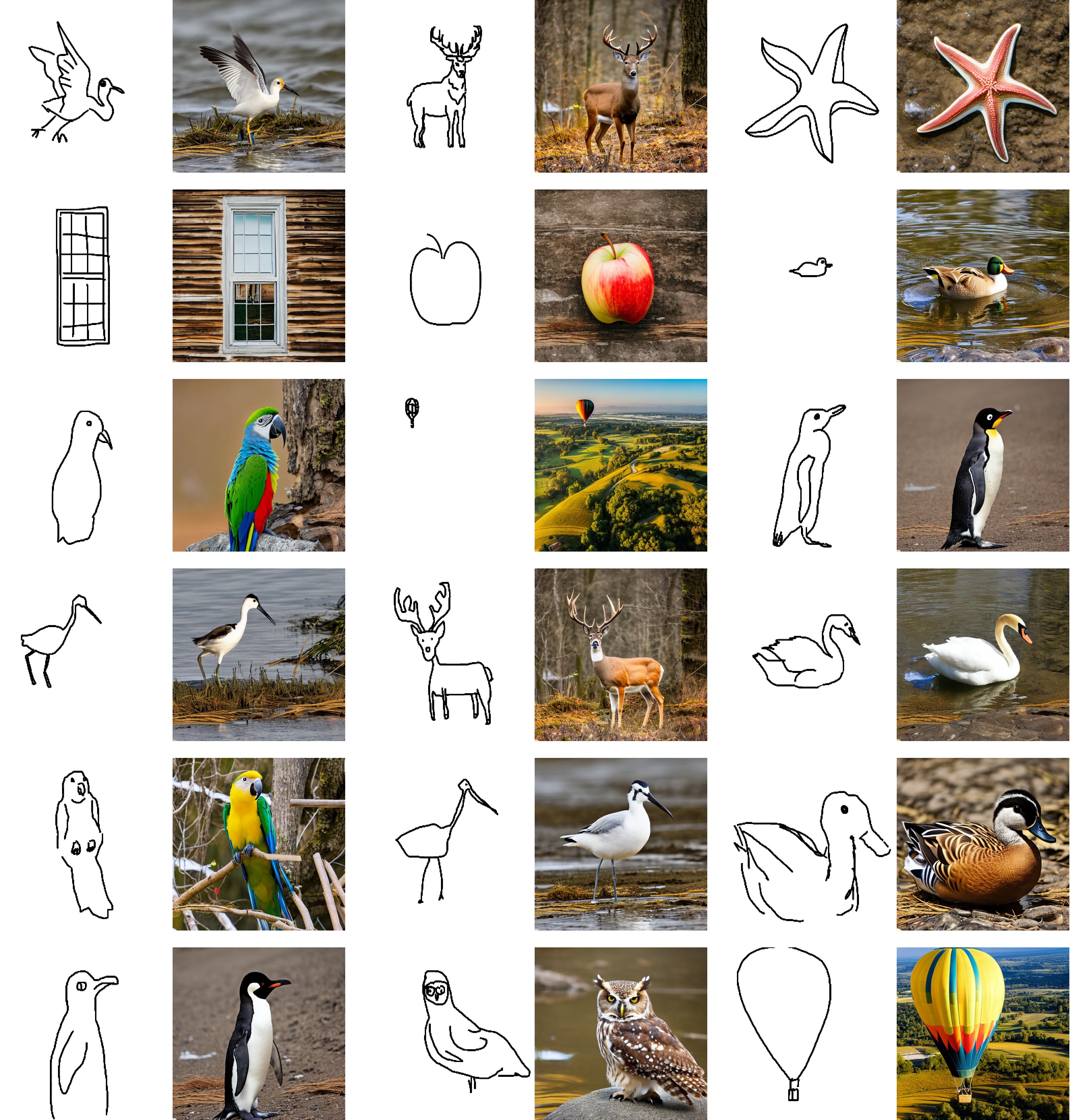}
    \vspace{-0.6cm}
    \caption{Images generated by our method.}
    \label{fig:qual_5}
\end{figure}

\begin{figure}[!htbp]
    \centering
    \includegraphics[width=1\linewidth]{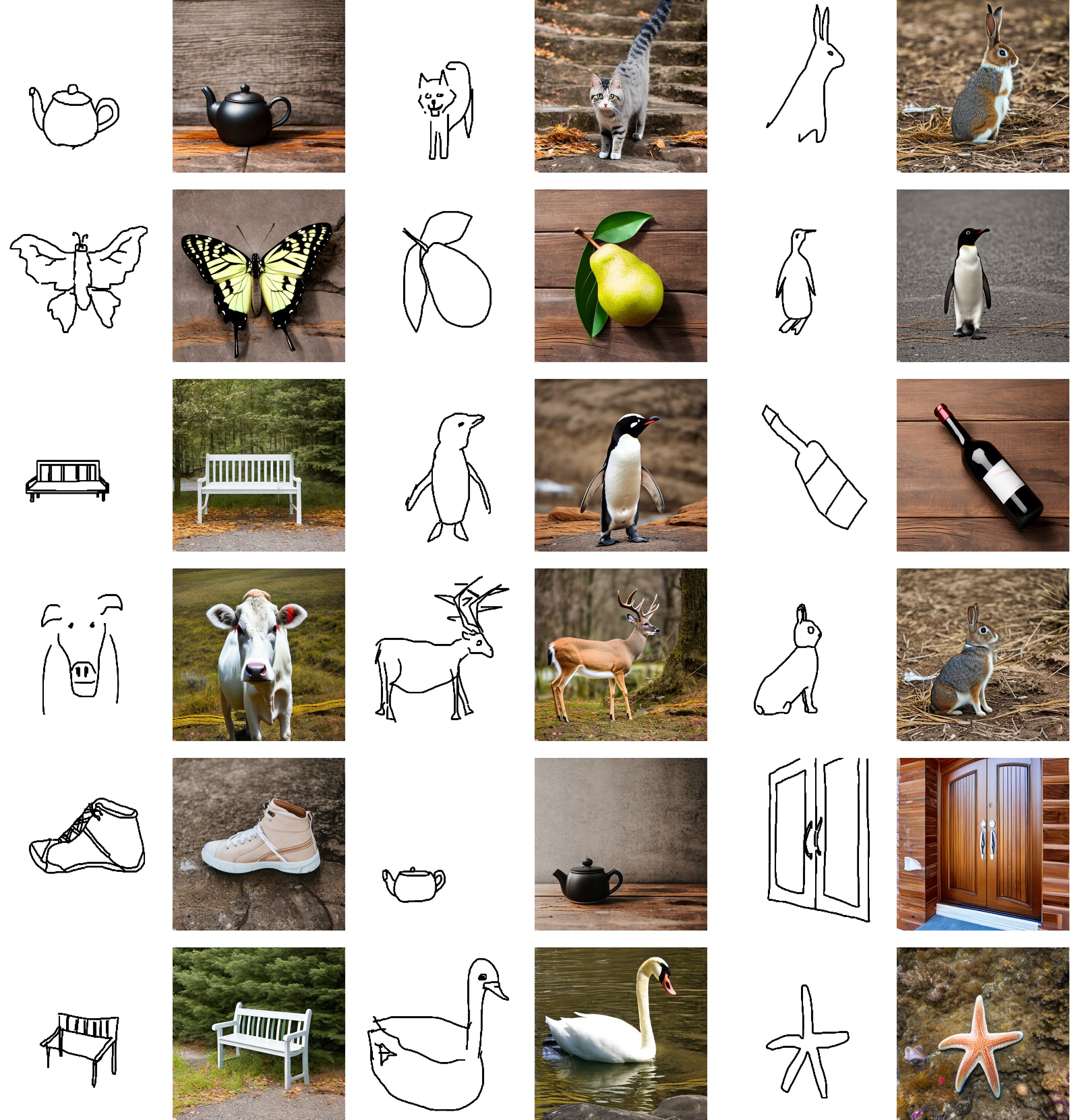}
    \vspace{-0.6cm}
    \caption{Images generated by our method.}
    \label{fig:qual_6}
\end{figure}

\vspace{-0.2cm}
\cref{fig:comp_1} delineates qualitative comparison of our method with pix2pix \cite{isola2017image}, CycleGAN \cite{zhu2017unpaired}, ControlNet \cite{zhang2023adding}, T2I-Adapter \cite{mou2023t2i}, and PITI \cite{wang2022pretraining}. Whereas, Fig.\ \ref{fig:qual_1}-\ref{fig:qual_6} shows additional results generated by our method.

\vspace{-0.cm}
\vspace{-0.3cm}

\vspace{-0.3cm}
\vspace{-0.3cm}

\vspace{-0.3cm}
\vspace{-0.3cm}

\vspace{-0.3cm}
\vspace{-0.3cm}

\vspace{-0.3cm}
\vspace{-0.3cm}

\vspace{-0.3cm}
\vspace{-0.3cm}

\vspace{-0.3cm}
\vspace{-0.3cm}
\clearpage

\section*{{B. Results on Out-of-distribution Sketches}} 
\begin{figure}[!htbp]
    \vspace{-2mm}
    \centering
    \includegraphics[width=\linewidth]{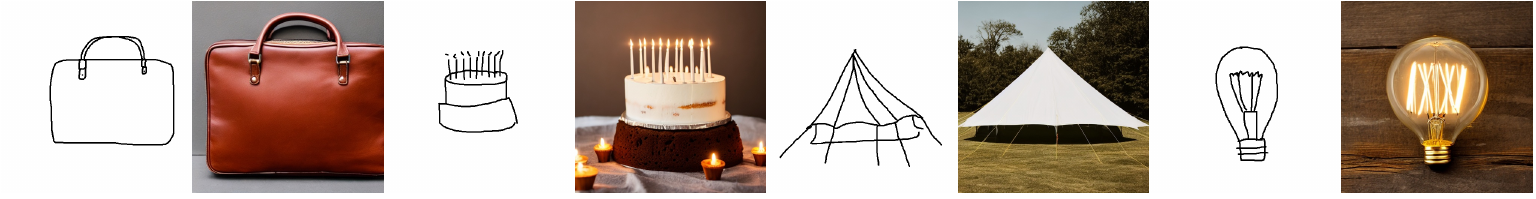}
    \caption{Results on out-of-distribution sketches.}
    \label{fig:ood}
    \vspace{-8mm}
\end{figure}

Keeping the pre-trained diffusion model \textit{frozen}, we fully leverage its generalisation potential. We posit that our design enables \text{out-of-distribution generalisation}. \cref{fig:ood} shows a few sketches, absent in Sketchy \cite{sangkloy2016sketchy}.

\end{document}